\documentclass{article}

\usepackage{microtype}
\usepackage[pdftex]{graphicx}
\usepackage{subfigure}
\usepackage{booktabs} 

\usepackage{hyperref}

\usepackage[T1]{fontenc}
\usepackage{lmodern}

\usepackage{amsmath}	
\usepackage{amsfonts}
\usepackage{bbm}        
\usepackage{amssymb}
\usepackage{theorem}

\newtheorem{theo}{Theorem}[section]

\newcommand{\R}{\mathbb{R}}

\DeclareMathOperator{\sgn}{sgn}

\title{Implicit Regularization in Over-parameterized Neural Networks}

\author{Masayoshi Kubo\thanks{Graduate School of Informatics, Kyoto University, kubo@i.kyoto-u.ac.jp}~~~~~
Ryotaro Banno\thanks{School of Informatics and Mathematical Science, Faculty of Engineering, Kyoto University}~~~~~
Hidetaka Manabe\footnotemark[2]~~~~~ Masataka Minoji\footnotemark[2]}

\date{\today}

\begin{document}
\maketitle

\begin{abstract}
Over-parameterized neural networks generalize well in practice
without any explicit regularization. 
Although it has not been proven yet, empirical evidence suggests
that implicit regularization plays a crucial role in deep learning
and prevents the network from overfitting.
In this work, we introduce the \textbf{gradient gap deviation} and
the \textbf{gradient deflection} as statistical measures corresponding to
the network curvature and the Hessian matrix to analyze variations of
network derivatives with respect to input parameters,
and investigate how implicit regularization works in ReLU neural networks
from both theoretical and empirical perspectives.
Our result reveals that the network output between each pair of
input samples is properly controlled by random initialization
and stochastic gradient descent to keep interpolating between
samples almost straight, which results in low complexity of
over-parameterized neural networks.
\end{abstract}

\section{Introduction}
\label{Introduction}
Deep neural networks (DNNs) have achieved high performance in
application domains such as computer vision,
natural language processing and speech recognition.
%
It is widely known that neural networks (NNs),
which are very often used in the over-parameterized regime,
generalize well without overfitting
\cite{neyshabur2014search, zhang2016understanding}.
However, the reason is not clear yet,
and related questions remain largely open
\cite{poggio2018theory,li2018learning}.
%
Recent works have shown that 
with over-parameterization and random initialization, 
stochastic gradient descent (SGD) can find the global minima of NNs
in polynomial time
\cite{du2018gradient,allen2018convergencet},
but trained network behavior in input data except for training samples
is still not well-understood, and seems to be relevant to
generalization in deep learning
\cite{wu2017towards}.

%
%
It has been known that a NN with i.i.d.\ random parameters
can be equivalent to a Gaussian process,
in the limit of infinite network width
\cite{Neal1994phdthesis,lee2017deep,matthews2018gaussian}.
%
NNs often have significantly
more parameters than samples in practice, and
increasing the number of parameters can lead to a decrease
in generalization error
\cite{balduzzi2017shattered, bartlett2017spectrally}.
%
It has been shown that the expressive power of NNs
grows exponentially with depth
for example 
by investigation into the number of linear regions
\cite{montufar2014number,telgarsky2016benefits},
analysis using Riemannian geometry and mean field theory
\cite{poole2016exponential},
and study using tensor decomposition
\cite{cohen2016expressive}.
%
If the enormous expressive power is not controlled,
deep learning architectures with large capacity
will not generalize properly.
However, these large NNs often generalize well
in practice, despite the lack of explicit regularization.
%
It is considered that implicit regularization plays a
crucial role in deep learning and prevents the NN from overfitting
\cite{zhang2016understanding,soudry2018implicit,li2018learning}.

\begin{figure*}[t]
 \begin{center}
  \begin{tabular}{c}
  
   \begin{minipage}{0.48\hsize}
    \begin{center}
     \includegraphics[width=7.5cm]{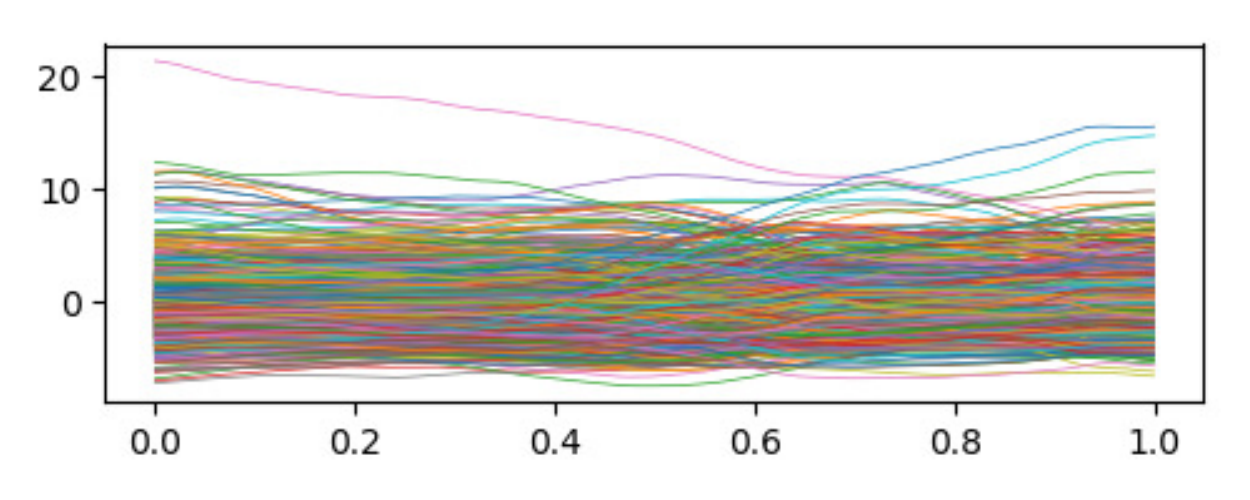}
     \hspace{0.2cm} (a) output on linear input path
    \end{center}
   \end{minipage}

   \begin{minipage}{0.48\hsize}
    \begin{center}
     \includegraphics[width=7.5cm]{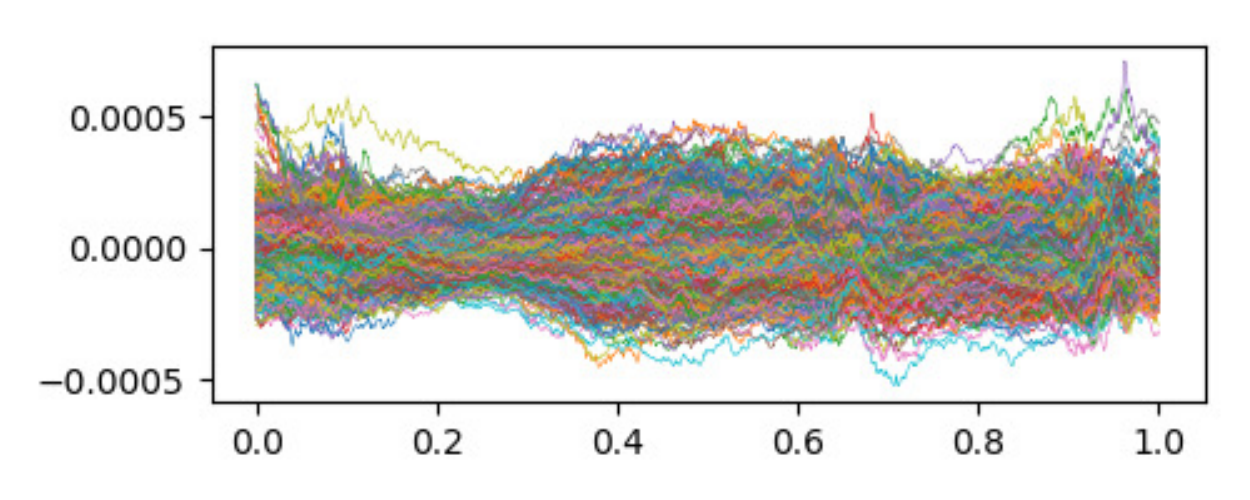}
     \hspace{0.2cm} (b) gradient on linear input path
    \end{center}
   \end{minipage}

  \end{tabular}
  \caption{ResNet-152 trained on ImageNet dataset \cite{he2016deep}:
  (a) The 1000 output components change as the input sweeps along
  the linearly interpolating path between
  a pair of samples in ImageNet. (b) The 1000 gradient components change
  as the input sweeps along the same path.}
  \label{fig:sample_path}
 \end{center}
\end{figure*}

%
%
In order to estimate variations of NN derivatives with respect to
input parameters, we define the \textbf{gradient gap deviation} and
the \textbf{gradient deflection},
and investigate how implicit regularization works
in ReLU activated neural networks (ReLU NNs)
from both theoretical and experimental perspectives.
%
%
Our result reveals that although the NN derivatives
between each pair of input samples seem to change almost randomly,
ReLU NNs interpolate almost linearly between the samples
because of small variations of their derivatives
(See Figure \ref{fig:sample_path}).
%
In other words, we show that the NN output between the samples
is properly controlled by weight initialization and SGD
to keep connecting the samples almost straight,
which results in low complexity of over-parameterized NNs.
This result is consistent with the theoretical
analysis and the experiments
\cite{nagarajan2017generalization, arora2018stronger,li2018learning}.

%
%
%
%
%
%

One key challenge in the behavior of NNs is that
the corresponding output functions and training objectives
are non-smooth due to the ReLU activation.
%
%
Whereas calculation of the network curvature and the Hessian matrix
w.r.t.\ input variables are necessary so as to make the evaluation of
such geometric complexity of ReLU NNs between each pair of input samples,
it is difficult to estimate them because the second order derivatives
cannot be defined as functions in $L^{\infty}$.
Even though the input changes linearly,
the ReLU NN output changes piecewise linearly
and has many non-differentiable points
(break points)\footnote{Strictly speaking, the NN output does not have gradient everywhere due to the non-smoothness of ReLU.}.
%
%
%
~~\\
\noindent
According to \cite{allen2018learning}:
\begin{quote}
 One may naively think that since a ReLU network is infinite-order
 differentiable everywhere except a measure zero set, so we can safely
 ignore the Hessian issue and proceed by pretending that the Hessian of
 ReLU is always zero. \underline{This intuition is very wrong}. (p.15)
\end{quote}
%
All information about network variations is ingeniously encoded into
the set of the break points caused by activation of ReLU, which
has measure zero.
%
We should analyze the behavior of the network output in an open neighborhood of
this zero measure set to decode its global geometric information.
%
In order to define alternatives to
the network curvature and the Hessian matrix,
it is necessary to characterize \textbf{gradient gaps},
that is, the slope changes at break points.
%
Accordingly, we model the NN derivative (NN gradient) as
a \textbf{random walk bridge} \cite{liggett1968invariance}
on a linearly interpolating path between each pair of input samples
(\textit{linear input path})
to express quantitatively the global variations of NNs.
%
By leveraging the random walk bridge,
we investigate how the NN gradient changes
as the input sweeps along the linear input path.

In the same line, the work in \cite{balduzzi2017shattered}
introduces a random walk for the gradient of a single hidden
layer network to show the gradient converges to Brownian motion
in the limit of infinite network width.
%
The work in \cite{raghu2016expressive,novak2018sensitivity}
analyzes trajectory length and the number of linear regions,
which measure how the NN output (not the NN gradient) changes
as the input sweeps along a one-dimensional path,
and find that the trajectory length grows exponentially with depth.
%
Comparing with these results,
we present theoretical estimates of the gradient gap deviation
and numerical calculations of the gradient deflection,
and evaluate random variations of NN gradients statistically
to study the effect of implicit regularization by 
random initialization and SGD.
%
%
These estimates depend only on
the feedforward map from input to output,
thus can be applied to a wide range of standard architectures
(e.g., VGG \cite{simonyan2014very} and ResNet \cite{he2016deep}).\\


\noindent
\textbf{Main Contributions}
\begin{itemize}
 \item 
       We model a ReLU NN gradient w.r.t. input variables 
       on a \textit{linear input path} as a \textbf{random walk bridge},
       and define the \textbf{gradient gap deviation}
       \eqref{gradient gap deviation},
       that is, the standard deviation of NN gradient gaps between samples.

 \item 
       We define the \textbf{gradient deflection} \eqref{gradient deflection}
       as the amount of global variations of a ReLU NN between samples.
       Our result shows that the standard network models (MLP, VGG and ResNet)
       on the standard datasets (MNIST, CIFAR10 and CIFAR100) 
       have small gradient deflection between samples 
       (i.e., they interpolate almost linearly between samples),
       and also reveals that one of the mechanisms of implicit
       regularization by SGD is to keep gradient gaps to be small.

 \item 
       We show that for the standard DNN models, the difference
       between the gradient gap deviation and the gradient deflection
       is relatively small,
       that is, the NN gradient is probably close to a random walk bridge
       on the linear input path.

 \item 
       We also estimate the NN output itself and experimentally
       investigate the difference between the output variations on the
       linear input path and the mean margin at each pair of samples,
       which relate to the observed performance.
\end{itemize}

\noindent
\textbf{Notation.} We use $[n]$ to denote $\{1,2, \ldots ,n\}$.
Let us denote the Euclidean norms of vectors $v$ by $\| v \|_2$, and
the spectral norms of matrices $M$ by $\|M\|_2$.
Let us denote the indicator function for event $E$ by $\mathbbm{1}_{E}$.

\section{Preliminaries}


Consider an $L$-layer fully-connected feedforward NN
with $m$ units in each hidden layer.
%
%
An activation function ReLU is given by $\phi(x) = \max\{0, x\}$,
and for a vector $v \in \R^m$  $(v = (v_1, \ldots ,v_m))$,
we define $\phi(v) := (\phi(v_1), \ldots , \phi(v_m))$.
The activation $h_{l}$ of hidden units and
the output $g_{l}$ of each layer $l \in [L-1]$ are given by

\begin{equation}
 \label{mlp}
  \begin{cases}
   g_{l}~ = W_l h_{l-1},~~~~~~~~~~~~~~~~~~~  &   l \in [L-1],
   ~~~~~~~~~~~~~~~~~~~~~~~~~~~~~~~~\\
   h_{l}~ = \phi(g_{l}) = \phi(W_l h_{l-1}), &    l \in [L-1],
   ~~~~~~~~~~~~~~~~~~~~~~~~~~~~~~~~\\
   f(x)   = W_L h_{L-1},~~~~~~~~~~~~~~~~~~~  & ~~
\end{cases}
\end{equation}
where $h_0 := x \in \R^{d}$ is the input to the network,
~$W_1 \in \R^{m \times d}$,~$W_l \in \R^{m \times m}$
~$(l \in \{2,3,\ldots ,L-1 \})$, and $W_L \in \R^{c \times m}$
are the weight matrices.
For the input $x$ and the weight matrices $W := (W_1, W_2, \ldots , W_L)$,
the NN output $f(x) \in \R^c$ is also denoted by $f(W,x)$.
%
Let us define indicator matrix $G_l(x)$,
the diagonal elements of which are activation patterns ($0$ or $1$)
at each layer  $l \in [L-1]$, as follows:
For $i,j \in [m]$,
\begin{equation}\label{indicator}
 (G_l(x))_{ij} :=
  \begin{cases}
   0                              & i \neq j \\
   \mathbbm{1}_{(g_l(x))_i \ge 0} & i = j
  \end{cases}
\end{equation}
Since the ReLU activation is positive homogeneous,
we obtain the following equality:
\begin{equation}
 \label{activation}
 \phi(W_l h_{l-1}(x)) = G_l(x) \cdot (W_l h_{l-1}(x)).
\end{equation}
For simplicity, we denote by $G(x)$ the set of indicator matrices
$\{G_l(x) \}_{l=1}^{L-1}$.
We assume the following weight initialization  \cite{he2015delving}:\\
\begin{equation}
 \label{initialization}
  \begin{cases}
   (W_1)_{ij}  \sim \mathcal{N}(0,2/d)~~~~
   &\forall i,j \in [m] \times [d],     \\
   (W_l)_{ij}  \sim \mathcal{N}(0,2/m)~~~
   &\forall i,j \in [m] \times [m], ~~\forall l \in \{2,3,\ldots ,L-1 \}, \\
   (W_L)_{ij}  \sim \mathcal{N}(0,2/m)~~~
   &\forall i,j \in [c] \times [m].
  \end{cases}
\end{equation}

\section{Random Walk Bridge for ReLU NNs}

In this section, we define
the \textbf{gradient gap deviation} and
the \textbf{gradient deflection} on each NN,
and show their theoretical estimates and experimental studies.
%
We evaluate the estimates on three canonical machine learning datasets:
MNIST \cite{lecun1998gradient},
CIFAR10, and CIFAR100 \cite{krizhevsky2009learning} in our experiments.
\subsection{Fully-connected Feedforword NNs (MLPs)}

Now we consider a fully-connected feedforward NN (MLP)
with ReLU activation for $c$-classes classification, and
show that the NN gradient between samples can be regarded as
a random walk bridge.

We denote by $\{ x_i \}_{i=1}^{n}$ and $\{ y_i \}_{i=1}^{n}$ 
respectively the training inputs and the labels,
where $x_i \in \R^d$ and $y_i \in \R^c$.
%
For weight matrices $W$ and an input $x$, 
we denote by $f(W,x)$ the NN output \eqref{mlp}.
The target function $L(W)$ in learning is the cross entropy loss over
the softmax defined as follows:
\[
 L(W) := - \frac{1}{n} \sum_{i=1}^{n}
 \log\left(
 \frac{\exp(f_{y_i}(W,x_i))}{\sum_{j=1}^{c} \exp(f_{j}(W,x_i)) } \right)
\]

We randomly choose $X_0, X_1 \in \R^d$  $(X_0 \neq X_1)$
from training data or test data, and
denote a direction vector by $v := X_1 - X_0$.
We define a linear interpolation between
$X_0$ and $X_1$ by $X(t) := (1-t) \cdot X_0 + t \cdot X_1$  $(t \in [0,1])$,
which is called the \textit{linear input path}.
We estimate the variation of the NN output pattern $f(W,X(t))$
on the linear input path $X(t)$  $(t \in [0,1])$.
%
Since the number of classes in the dataset is $c$,
$f(W,x)$ is a $c$-dimensional vector
$\{ f_1(W,x), \ldots, f_{c}(W,x) \}$.
%
%
We will link the directional derivative of each component function $f_j(W,x)$ 
to a \textbf{random walk bridge}.
%
%
For $j \in [c]$, we define the NN output $u(t;X_0, X_1)$ between two points
$X_0, X_1$  $(X_0 \neq X_1)$ as follows:
\[
u(t;X_0, X_1)~~ (= u_j (t;X_0, X_1))~ := f_j(W,X(t)).
\]
For simplicity, we denote by $u(t)$ the NN output $u(t;X_0, X_1)$.
%
We define the \textbf{NN gradient} between $X_0$ and $X_1$
as follows\footnote{It can be defined as a function $\in L^{\infty}(0,1)$}:
\[
\nabla_v u (t) ~~(= \nabla_v u (t;X_0, X_1))~
:= \lim_{h\to 0} \frac{u(t + h) - u(t)}{h\|v\|}.
\]
%
Here NN gradient is normalized such that it is independent of the length of
a direction vector $v$.
%
In particular, if $\|v\| = 1$ then $\nabla_v u (t)$ is equal to
an ordinary derivative $\frac{d}{dt} u(t)$.

Since $u(t)$ is a piecewise linear function due to ReLU
\cite{balduzzi2017shattered,raghu2016expressive},
we can define \textbf{nodes} of $u(t)$
(i.e., break points at which the sign of the
input to ReLU $\phi$ is switched)
as $0 < t_1 < t_2 < \cdots < t_{\mathcal K} < 1$.
%
Here, $\mathcal{K}$ is the number of times $u(t)$ breaks
in the interval $[0,1]$.
%
For notational simplicity, we denote both ends of the interval $[0, 1]$
by $t_0 = 0$ and $t_{\mathcal{K}+1} = 1$ respectively.
%
The number of nodes $\mathcal{K}$ is equivalent to how many times
the input linear path $X(t)$ passes through other linear regions.
The total number of linear regions of the input space
is studied as a measure to evaluate
expressivity and approximating ability of NNs
\cite{montufar2014number,raghu2016expressive,telgarsky2016benefits}.
%
For a single hidden layer MLP with ReLU activation (i.e., $L=2$),
we can estimate the number of nodes $\mathcal{K}$ as follows.
This estimate can be proved by using random initialization, 
over-parameterization and input randomness.
%
Moreover, since over-parameterization and random initialization
jointly restrict every weight vector to be close to
its initialization for all iterations
\cite{du2018gradient,allen2018convergencet},
it follows that this estimate changes little even after training.
 \begin{theo}[Estimate of Number of Nodes]
  \label{NumberNodes}
 Consider a two-layer fully-connected NN
 with $m$ rectified linear units and $d$-dimensional input.
 Let $X_0$, $X_1 \in \R^d$ be random vectors with i.i.d.\ entries 
 $(X_0)_{i}, (X_1)_{i} \sim \mathcal{N}(0,1)$,
 and let $W_1 \in \R^{m\times d}$ be a weight matrix of the first layer
 with i.i.d.\ entries $(W_1)_{i,j} \sim \mathcal{N}(0, 2/d)$.
 Then the number of nodes $\mathcal{K}$ in the interval $(X_0, X_1)$
 is in distribution identical to binomial distribution
 $\mathcal{B}(m, 1/2)$ with $m$ trials and $1/2$ success rate.
 \end{theo}
Our experimental results show that 
the mean of the number of nodes is approximately half of
the number of entire hidden units for an even deeper network.
%
%

%
For each interval $I_k := (t_k, t_{k+1})$ and for each layer $l$,
we denote by $G_l^{(k)} := G_l (X(t))$~ $(t \in I_k)$
the indicator matrix \eqref{indicator} on the linear input path.
We also abbreviate $\{ G_l^{(k)}\}_{l=1}^{L-1}$ as $G^{(k)}$.
The diagonal elements of $G_l^{(k)}$ show ReLU activation of layer $l$
in the interval $I_k$.
By definition, the indicator matrices $G^{(k)}$ $k \in [\mathcal{K}]$
are different from each other.
Since $G^{(k)}$ is constant in each interval $I_k$,
$u(t)$  $(t \in I_k)$ is a linear function and
$\nabla_v u (t)$  $(t \in I_k)$ is a constant function.
%
For $t \in I_k$  $(k \in [\mathcal{K}])$,
we use $R_k := \nabla_v u (t)$ to denote NN gradient.
For each node $t_k$, we define \textbf{NN gradient gap} as
$Y_k := R_k - R_{k-1}$  $(k \in [\mathcal{K}])$.
This implies that
\begin{equation}
 \label{nn_grad_gap}
  \begin{cases}
   \displaystyle
   R_k = \sum_{i=1}^{k} Y_i,                      ~~&~~ \\
   R_0 = \nabla_v u(0),  ~~R_{\mathcal{K}} = \nabla_v u(1).  ~~&~~
  \end{cases}
\end{equation}

We define a probability measure $P$ as distribution of NN gradient gaps
when choosing $X_0, X_1~~(X_0 \neq X_1)$ randomly,
and denote by $\sigma^2 := \int z^2 dP(z)$ the variance of $P$.
We assume that the mean of $P$ is zero, and
define $Z_1,Z_2,Z_3, \ldots$ as i.i.d.\ random variables
from the probability distribution $P$.
%
Then we model the NN gradient $\nabla_v u (t)$
as a \textbf{random walk bridge} $\{ S_k \}$ generated by $P$,
which satisfies the following:
\begin{equation}
 \label{bridge}
  \begin{cases}
   \displaystyle
   S_k = \sum_{i=1}^{k} Z_i,           ~~&~~ \\
   S_0 = \nabla_v u(0),  ~~S_{\mathcal{K}} = \nabla_v u(1). ~~&~~
  \end{cases}
\end{equation}
%
If gradient gaps $Y_1, \ldots , Y_{\mathcal{K}}$ are i.i.d.\ random variables,
then $\{R_k\}$ is the random walk bridge satisfying
the boundary conditions on $R_0$ and $R_{\mathcal{K}}$.
%
For a standard random walk,
the boundary value at one out of both ends is fixed.
For a random walk bridge,
the both end values are fixed,
which is represented by a conditional probability.
In this setting \eqref{bridge}, the following theorem evaluates
to what extent a random walk bridge $\{ S_k \}$ deviates from
a linear interpolation between the both ends.
\begin{theo}[Estimate of Gradient Gap Deviation]
 \label{GradientGapDeviation}
 Let $\{ S_k \}_{k=1}^{\mathcal{K}}$ be a random walk bridge \eqref{bridge}
 generated by a probability distribution $P$
 with mean $0$ and variance $\sigma^2$,
 and let $\{ T_k \}$ be a random walk bridge defined by
 $T_k := (S_k -S_0) - \frac{k}{\mathcal{K}} (S_{\mathcal{K}} -S_0)$.
 We define \textbf{gradient gap deviation} for distribution $P$
 as follows:
 \begin{equation}
  \label{gradient gap deviation}
   \textup{\textbf{(Gradient Gap Deviation)}}
   := \left\{ E\left[ T_k^2 \right] \right\}^{1/2}.
 \end{equation}
 Then the \textbf{gradient gap deviation} is equal to 
 $\sigma \sqrt{k(1- k/{\mathcal{K}})}~~(0\le k \le \mathcal{K})$.
\end{theo}

Using the gradient gap deviation, 
which attains the maximum value
$\displaystyle \frac{1}{2} \sigma \sqrt{\mathcal{K}}$
at the midpoint $(k = \frac{\mathcal{K}}{2})$,
we estimate stochastically
the difference between a random walk bridge and the linear interpolation
from one end to the other.

In contrast, we also define the difference $D(t;X_0,X_1)$
between the NN gradient on the linear input path $X(t)$ and
a linear interpolation of both end values of the NN gradient:
For the input samples $X_0$ and $X_1$~~$(X_0 \neq X_1)$,
\[
D(t;X_0,X_1) := \nabla_v u(t;X_0,X_1) - (\nabla_v u(1;X_0,X_1)
 - \nabla_v u(0;X_0,X_1)) t - \nabla_v u(0;X_0,X_1). 
\]
%
Suppose $X_0$ and $X_1$~~$(X_0 \neq X_1)$ are chosen uniformly
from a certain training or test dataset.
Then we define deflection of the NN gradient (\textbf{gradient deflection})
by the following expectation:
\begin{equation}
 \label{gradient deflection}
 \textbf{(Gradient Deflection)} :=
 \left\{ E_{X_0 \neq X_1}\left[ D(t;X_0,X_1)^2 \right] \right\}^{1/2}.
\end{equation}
%
Here the difference between 
the \textbf{gradient gap deviation} \eqref{gradient gap deviation} and 
the \textbf{gradient deflection} \eqref{gradient deflection}
is corresponding to the one between
the amount of change in the random walk bridge
and the network output change in the actual case.
%
Inspecting this difference enables us to statistically measure
the extent to which the NN changes contrivedly.
Then, since the distribution $P$ is determined by activation patterns of ReLU,
we obtain the following estimate:
\begin{theo}[Estimate of Gradient Gap]
 \label{GradientGap}
 Consider an $L$-layer fully-connected neural network
 with $m$ rectified linear units in each layer.
 If $\varepsilon \in (0, 1]$, 
 with probability at least $1- e^{- \Omega(m\varepsilon^2/L)}$
 over the randomness of $W$, then gradient gaps are in distribution
 almost identical to the combination of
 $\mathcal{N}(0, \frac{4}{md})$ and $\mathcal{N}(0, \frac{4}{m^2})$.
\end{theo}

\begin{figure*}[t]
 \begin{center}
  \begin{tabular}{c}
  
   \begin{minipage}{0.32\hsize}
    \begin{center}
     \includegraphics[width=5.0cm]{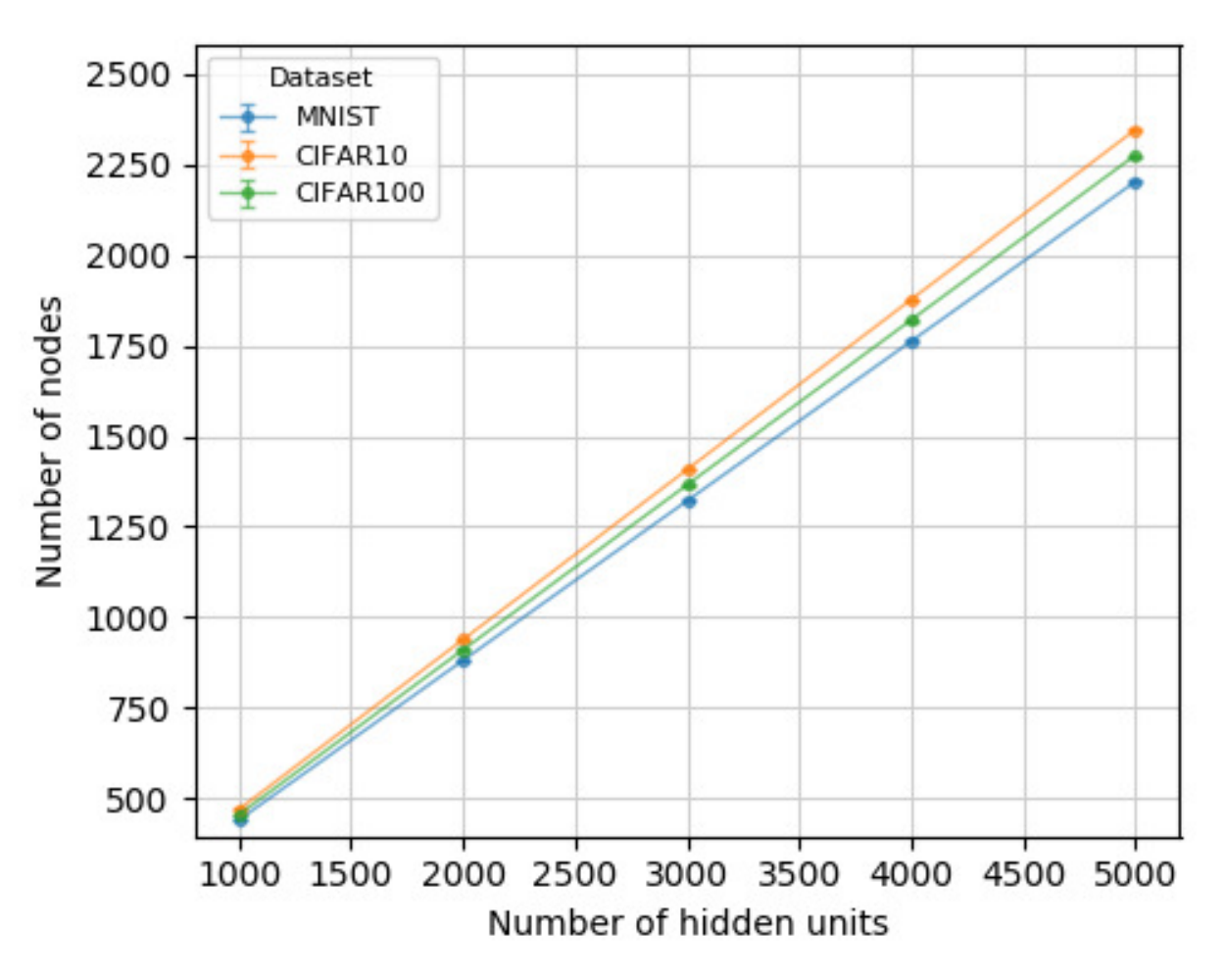}
     \hspace{0.2cm} (a) Number of nodes
    \end{center}
   \end{minipage}

   \begin{minipage}{0.32\hsize}
    \begin{center}
     \includegraphics[width=5.0cm]{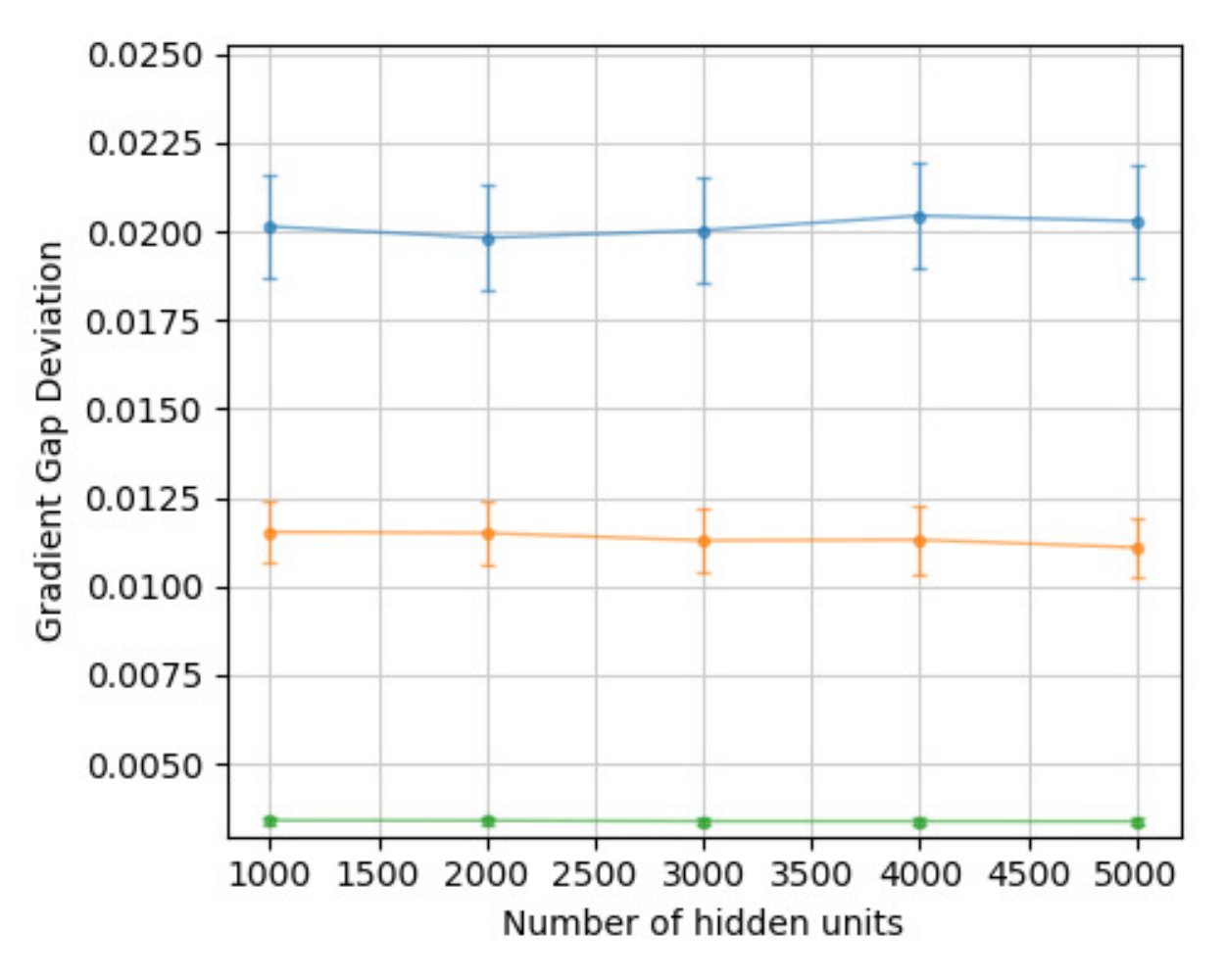}
     \hspace{0.2cm} (b) Gradient Gap Deviation
    \end{center}
   \end{minipage}

   \begin{minipage}{0.32\hsize}
    \begin{center}
     \includegraphics[width=5.0cm]{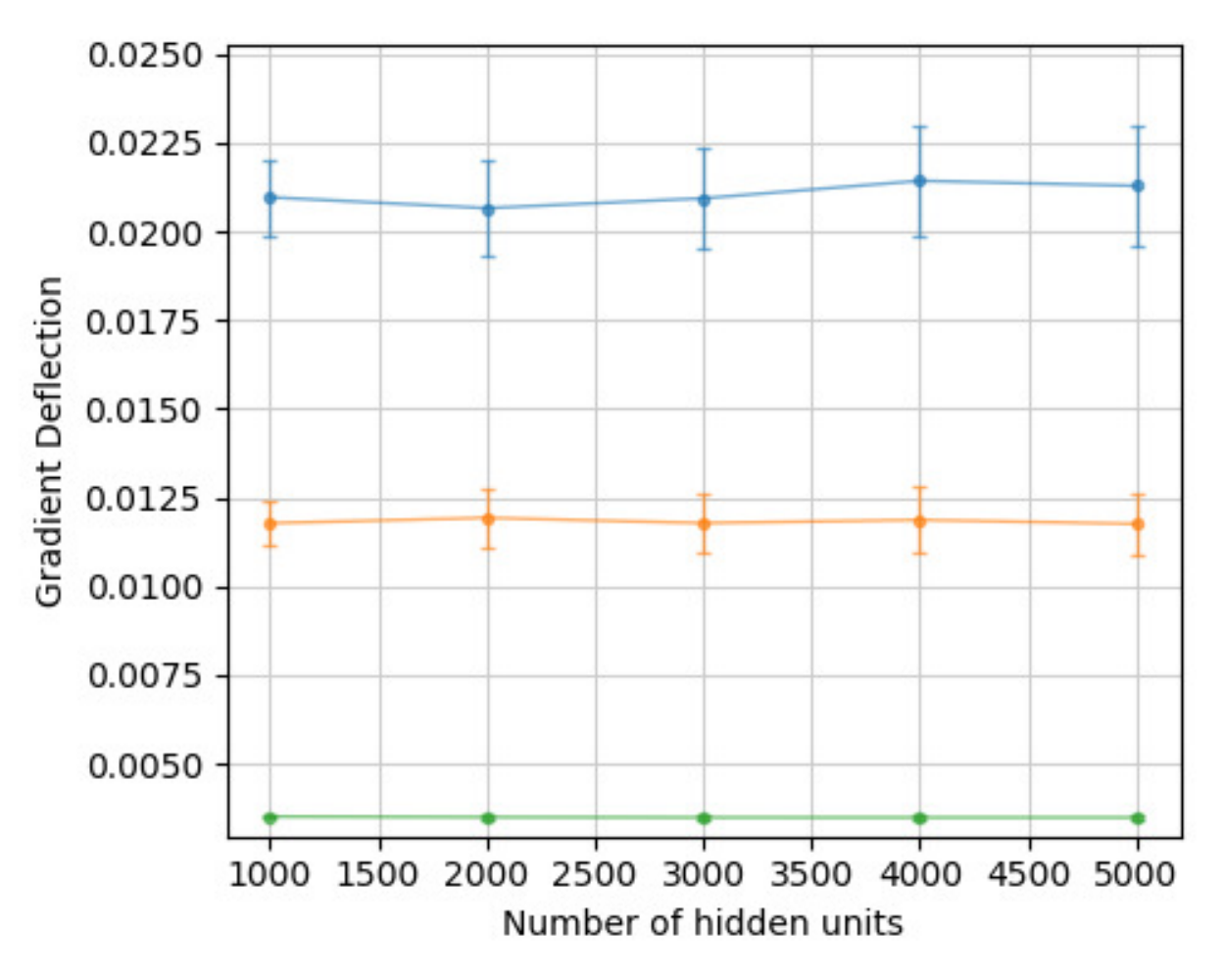}
     \hspace{0.2cm} (c) Gradient Deflection
    \end{center}
   \end{minipage}

  \end{tabular}
  \caption{The two-layer MLPs in the initial state on linearly
  interpolating paths between each pair of samples in each dataset (
  MNIST, CIFAR10 and CIFAR100). Plots are averaged over 100 networks
  for each of the different number of hidden units
  (a) The number of break points (nodes).
  (b) Gradient gap deviation (i.e.,the standard deviation of NN
  gradient gaps between samples.) (c) Gradient deflection
  (i.e.,the amount of global variation of the NN output between samples).}
  \label{fig:mlp_dataset}
 \end{center}
\end{figure*}
%
%
%

First we show experimental observations of the two-layer MLPs
in the initial state (without training).
Figure \ref{fig:mlp_dataset} (a) shows that the number of nodes $\mathcal{K}$
is approximately equal to its theoretical value $\frac{m}{2}$.
The values of gradient gap deviation and gradient deflection are small
$(\ll 1.0)$ in Figure \ref{fig:mlp_dataset} (b) and (c).
Even if the number of units $m$ increases, both values of gradient gap
deviation and gradient deflection hardly vary. 
In particular, the theoretical value of gradient gap deviation
(Theorem \ref{GradientGapDeviation})
for the two-layer MLPs in the initial state is as follows:
\[
 \frac{1}{2}\sigma \sqrt{\mathcal{K}} = 
 \frac{1}{2}\sqrt{\frac{4}{md}} \sqrt{\frac{m}{2}} = \frac{1}{\sqrt{d}}.
\]
where $m$ is the number of hidden units, and $d$ is the input dimensions.

~~\\
\noindent
\textbf{Normalization procedure for network output.}
If the variance of the output pattern $f(W,x)$ (i.e., $\|f(W,x)\|_2^2$)
changes by training or network architecture,
it is difficult to compare with other NN gradients $\nabla_v u (t)$.
Weights are initialized randomly so that the variance of the input
will not change in each layer \cite{he2015delving}.
It is known that much deeper NNs can be trained by intensifying
such effect of initialization \cite{xiao2018dynamical,zhang2019fixup}.
%
In this work, we adopt the following method, which is the same as
\cite{dinh2017sharp,bartlett2017spectrally,liao2018surprising}:
%
Without loss of accuracy in classification,
we can replace the output $f(W,x)$ with $f(W,x)/\alpha_{x}$,
where $\alpha_{x} >0$ is an arbitrary fixed number for each input $x$.
%
Therefore, we define each fixed number $\alpha_{x}$ for normalization of
random walk bridge as follows:
For each input $x$, 
define $\alpha_{x}$ as the $L^2$ norm of the output vector $f(W,x)$.
%
In other words, we normalize the NN output by letting
$\tilde{f}(W,x) := f(W,x)/\|f(W,x)\|_2 $.
%
Now using the normalized NN output $\tilde{f}(W,x)$
enables us to estimate random walk bridge
even if the weights change with SGD training steps.

~\\~~\\
\noindent
\textbf{Numerical finite difference approximations of derivatives.}
As the number of units or the number of layers increases
the number of nodes becomes large and the length of each
interval $I_k = (t_k, t_{k+1})$ becomes extremely short.
In particular, we cannot calculate precisely
the derivative of the output function $u(t)$ 
by subtracting in the short interval 
owing to \textit{cancellation of significant digits}
in GPU calculation using single precision floating point arithmetic.
Instead, we propose a method of replacing each ReLU activation layer with
indicator matrix \eqref{indicator},
and only need to calculate the product of the matrix
(See Appendix \eqref{product of matrix}).\\

\begin{figure*}[t]
 \begin{center}
  \begin{tabular}{c}
  
   \begin{minipage}{0.32\hsize}
    \begin{center}
     \includegraphics[width=5.0cm]{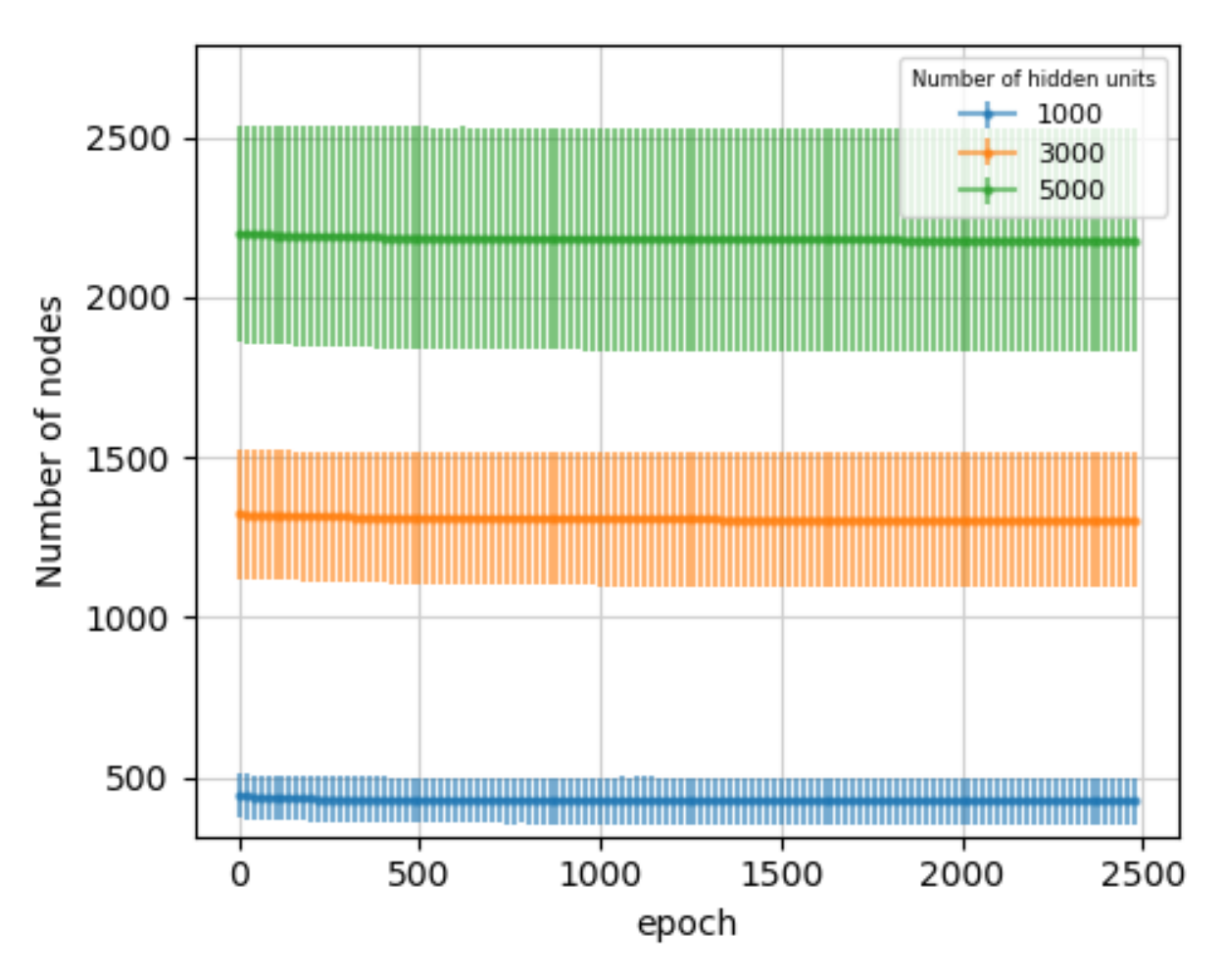}
     \hspace{0.2cm} (a) Number of nodes
    \end{center}
   \end{minipage}

   \begin{minipage}{0.32\hsize}
    \begin{center}
     \includegraphics[width=5.0cm]{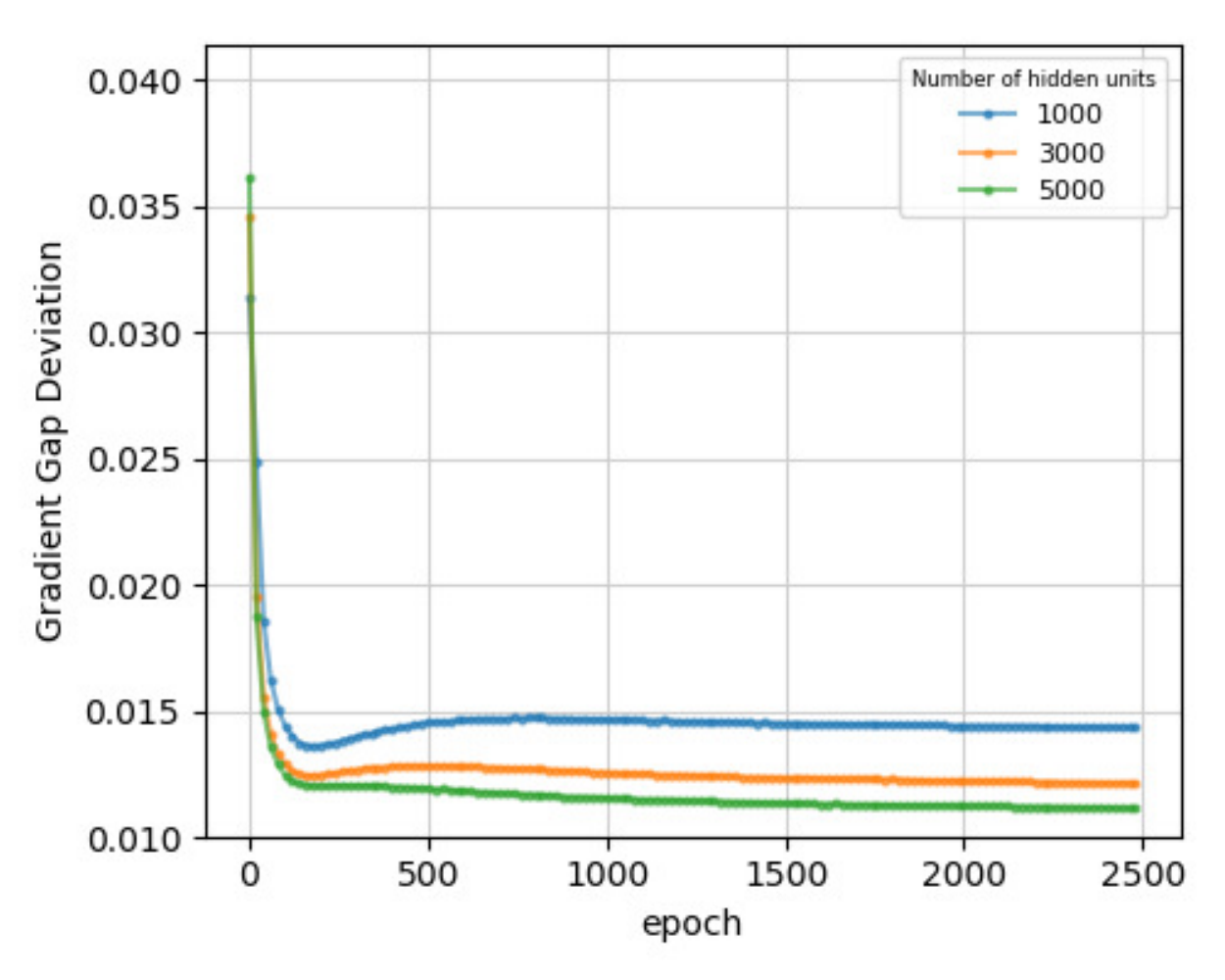}
     \hspace{0.2cm} (b) Gradient Gap Deviation
    \end{center}
   \end{minipage}

   \begin{minipage}{0.32\hsize}
    \begin{center}
     \includegraphics[width=5.0cm]{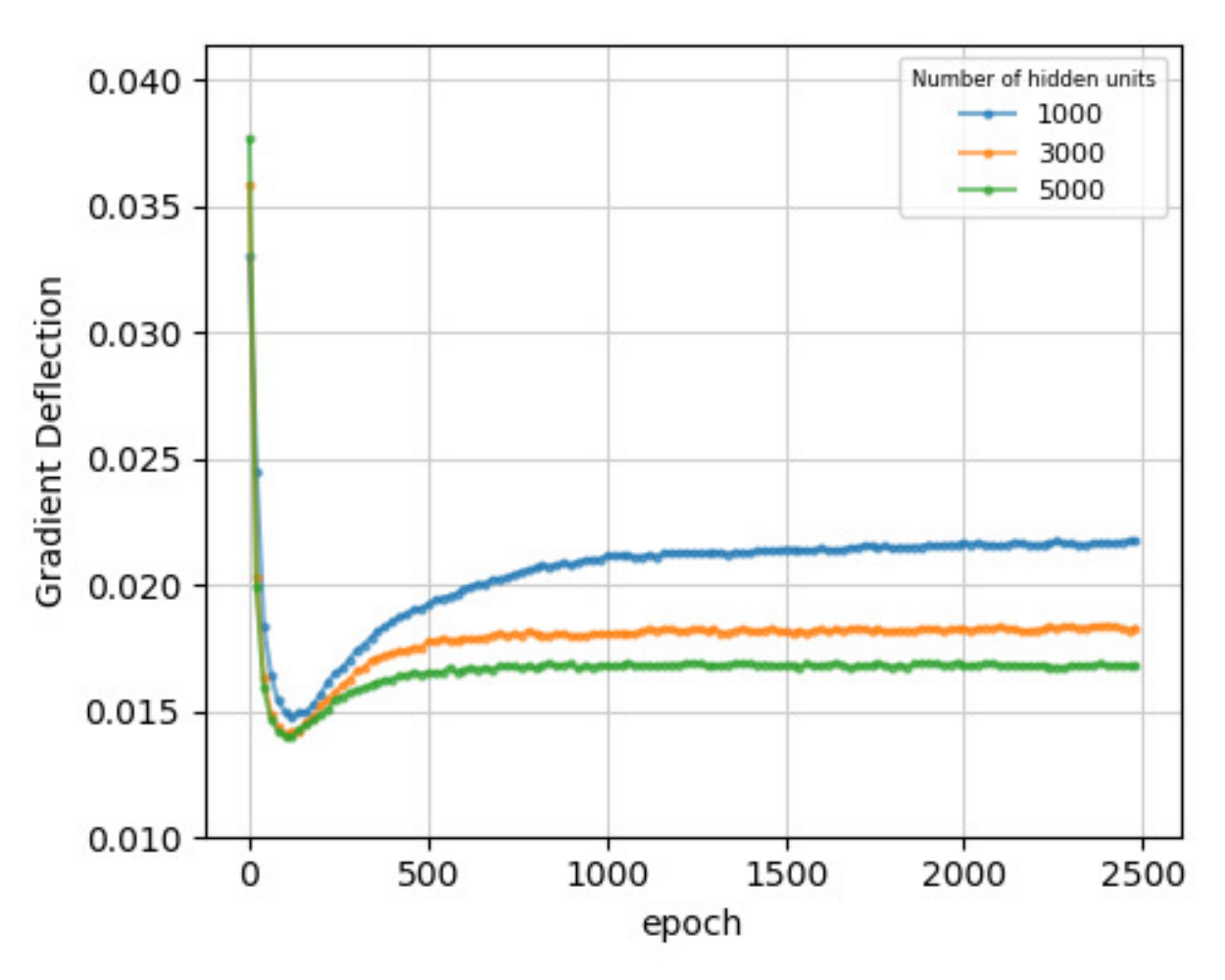}
     \hspace{0.2cm} (c) Gradient Deflection
    \end{center}
   \end{minipage}

  \end{tabular}
  \caption{The $2$-layer MLPs on linearly interpolating paths between each pair of samples changes with SGD training steps on MNIST.}
  \label{fig:mlp_sgd}
 \end{center}
\end{figure*}

Next we show that the complexity measures
(the number of nodes, the gradient gap deviation and the gradient deflection)
change with SGD training steps
for two-layer MLPs with different number of units.
%
We observe that the number of nodes converges relatively fast
and does not change significantly in Figure \ref{fig:mlp_sgd} (a).
%

%
%
%
%
%
%

We investigate how randomly an MLP changes between samples.
%
Let $S_k$ denote a random walk bridge \eqref{bridge}
generated by the probability distribution $P$.
Figure \ref{fig:mlp_sgd} (b) and (c) show
the gradient gap deviation of $S_k$ and the gradient deflection
(i.e., deviation of actual gradient change of MLP: $R_k$ ) at midpoint
$k = \left\lfloor \frac{\mathcal{K}}{2} \right\rfloor$
with SGD training steps.
Although in the above-mentioned figures we sample randomly 
a pair of $X_0, X_1$ from training data,
we obtain the same result even when we sample randomly them from test data.

\subsection{VGG and Residual Networks}
%
Now we show that for
VGG \cite{simonyan2014very} and residual network (ResNet)
\cite{he2016deep,he2016identity},
the NN gradient between samples can be regarded as a random walk bridge.
We also show experimental results of
the gradient gap deviation and the gradient deflection.

Let us denote the output of VGG or ResNet by $f(W,x)$ 
for weight matrices $W$ and an input sample $x$.
As with \S 3.1, we investigate the variation of 
the NN output $f(W,x)$ between two input samples
$X_0, X_1 \in \R^d  ~~(X_0 \neq X_1)$.
%
Let $X(t)$ be a linear interpolation $(1-t) \cdot X_0 + t \cdot X_1$,
and $v$ be a direction vector $X_1 - X_0$.
%
Then, by using the NN output $f(W,x)$,
we similarly define $u(t)$ and $\nabla_v u (t)$
between two points $X_0, X_1$ as follows:
\begin{align*}
 u(t)           & := f(W,X(t)), \\
 \nabla_v u (t) & :=
 \lim_{h\to 0} \frac{u(t + h) - u(t)}{h\|v\|}.
\end{align*}
%
Let $0 =t_0 < t_1< t_2< \cdots < t_{\mathcal{K}} < t_{\mathcal{K}+1} = 1$
be the nodes of a piecewise linear function $u(t)$.
%
The number of nodes $\mathcal{K}$ corresponds to
the number of linear regions intersecting the linear input path.
%
The number of linear regions of deep models (VGG and ResNet)
grows exponentially in $L$ and polynomially in $m$,
which is much faster than that of shallow models with $mL$ hidden units
\cite{montufar2014number,serra2017bounding}.
%
In contrast, our experimental results for VGG and ResNet show that 
the number of nodes is half of the number of entire hidden units or less.
%
%
The number of nodes increases linearly rather than exponentially
with the depth (the number of layers),
which is completely different from linear regions.
%
Thus, by Theorem \ref{GradientGapDeviation},
gradient gap deviation, which is the fluctuation of random walk bridge,
is relatively small.
%
Since in the over-parameterized setting,
SGD learns a NN close to the random initialization,
the number of nodes is proved to change little.

\begin{figure*}[t]
 \begin{center}
  \begin{tabular}{c}
  
   \begin{minipage}{0.32\hsize}
    \begin{center}
     \includegraphics[width=5.0cm]{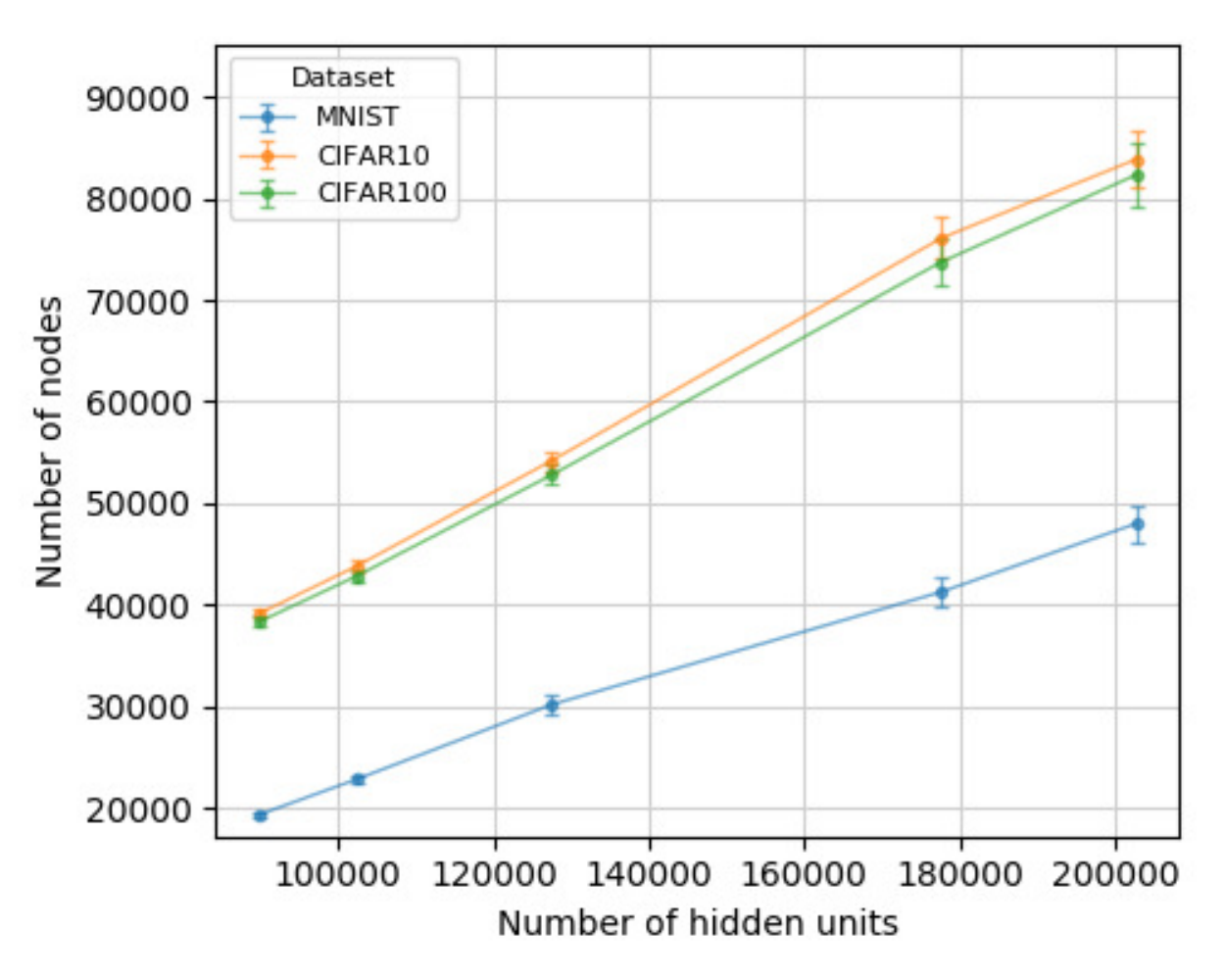}
     \hspace{1.0cm} (a) Number of nodes
    \end{center}
   \end{minipage}

   \begin{minipage}{0.32\hsize}
    \begin{center}
     \includegraphics[width=5.0cm]{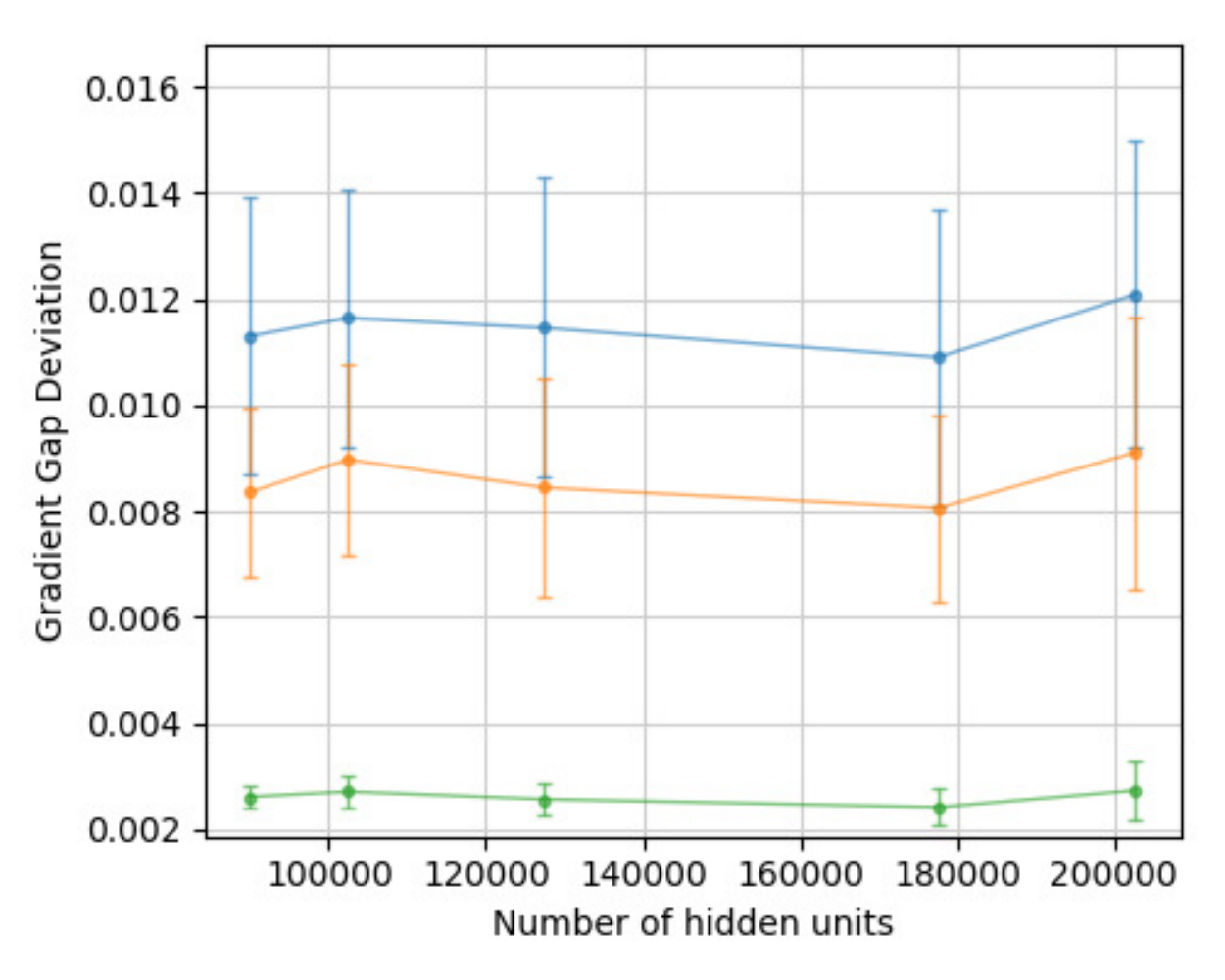}
     \hspace{1.0cm} (b) Gradient Gap Deviation
    \end{center}
   \end{minipage}

   \begin{minipage}{0.32\hsize}
    \begin{center}
     \includegraphics[width=5.0cm]{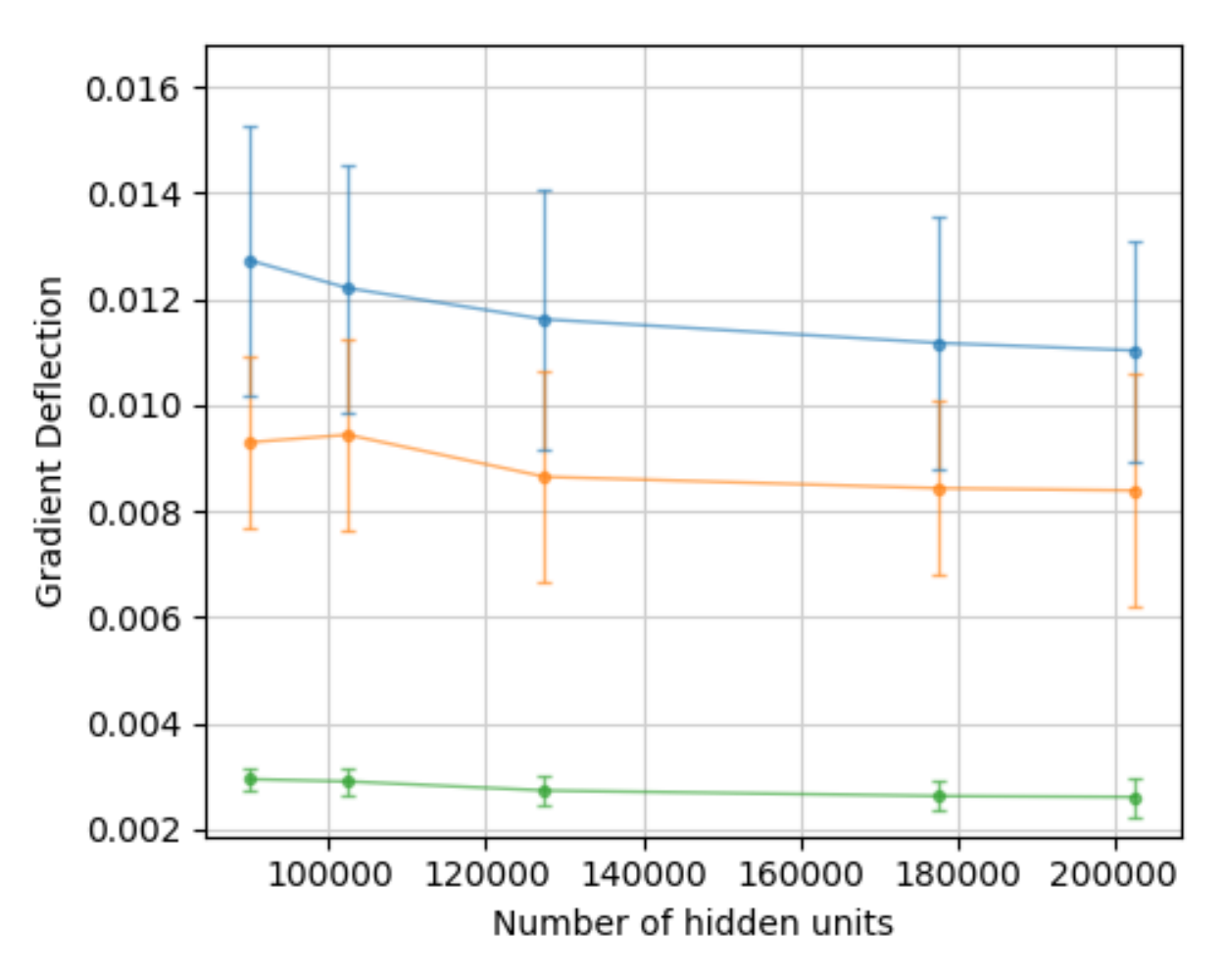}
     \hspace{1.0cm} (c) Gradient Deflection
    \end{center}
   \end{minipage}
   \\~~\\ 
   \begin{minipage}{0.32\hsize}
    \begin{center}
     \includegraphics[width=5.0cm]{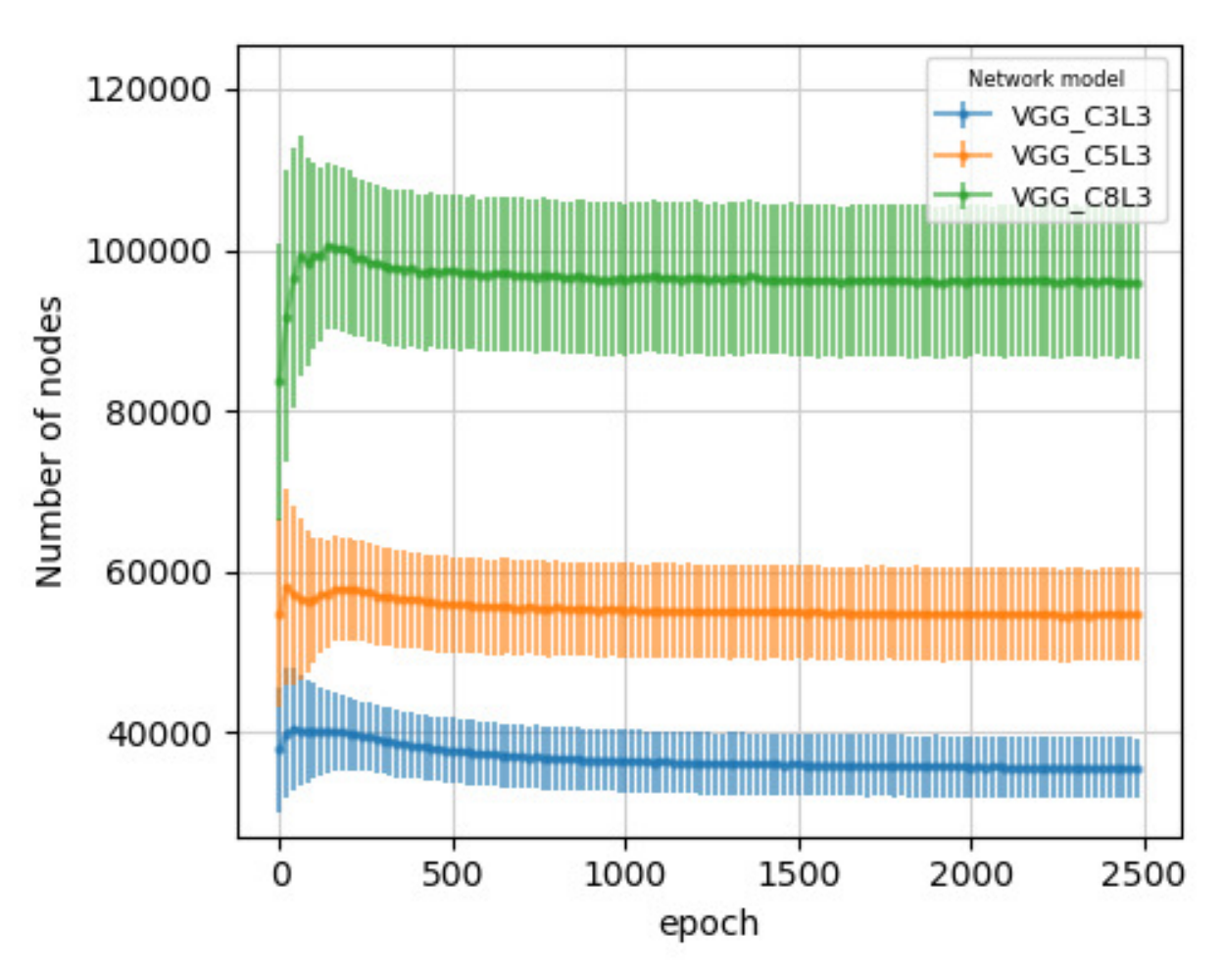}
     \hspace{1.0cm} (d) Number of nodes
    \end{center}
   \end{minipage}

   \begin{minipage}{0.32\hsize}
    \begin{center}
     \includegraphics[width=5.0cm]{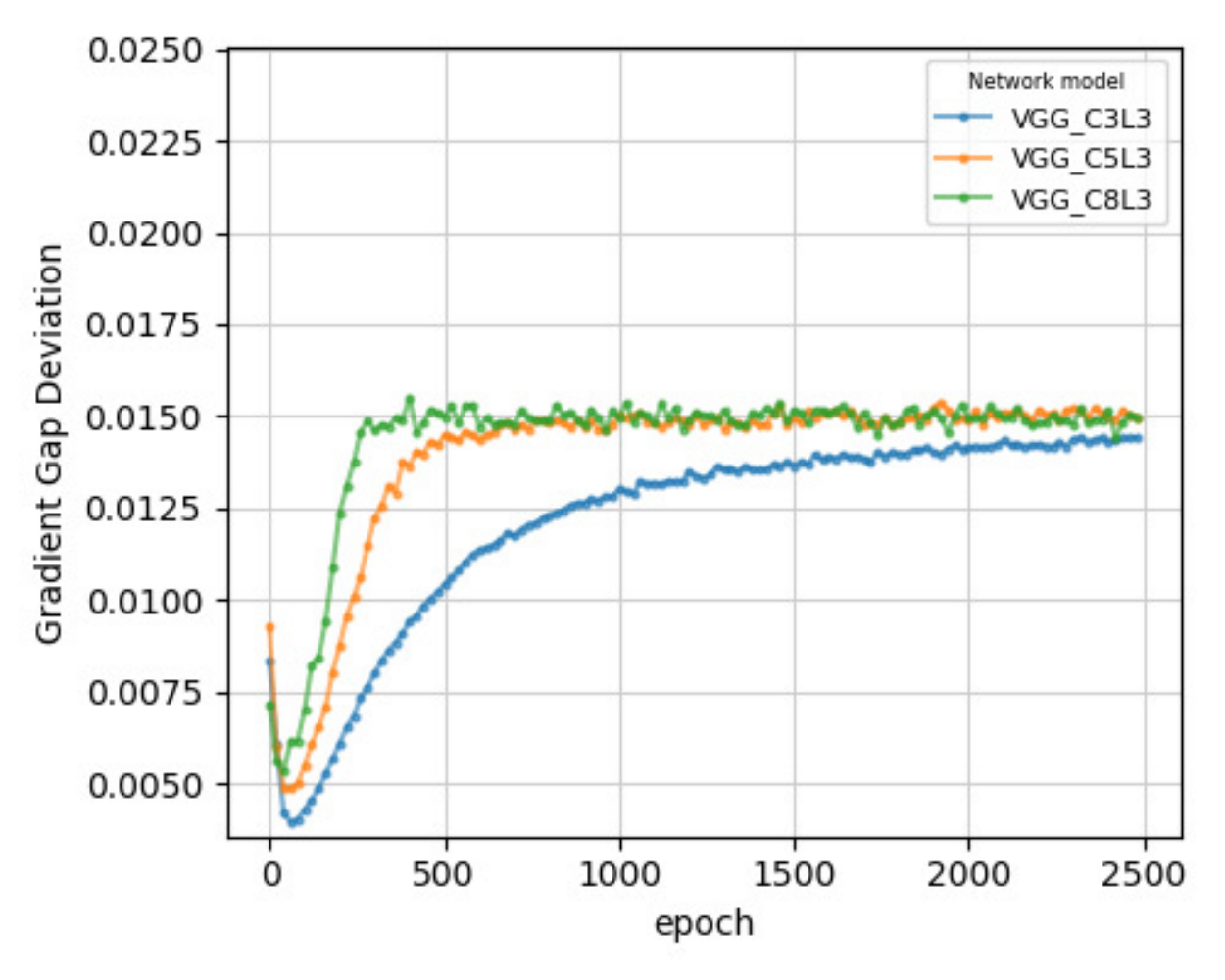}
     \hspace{1.0cm} (e) Gradient Gap Deviation
    \end{center}
   \end{minipage}

   \begin{minipage}{0.32\hsize}
    \begin{center}
     \includegraphics[width=5.0cm]{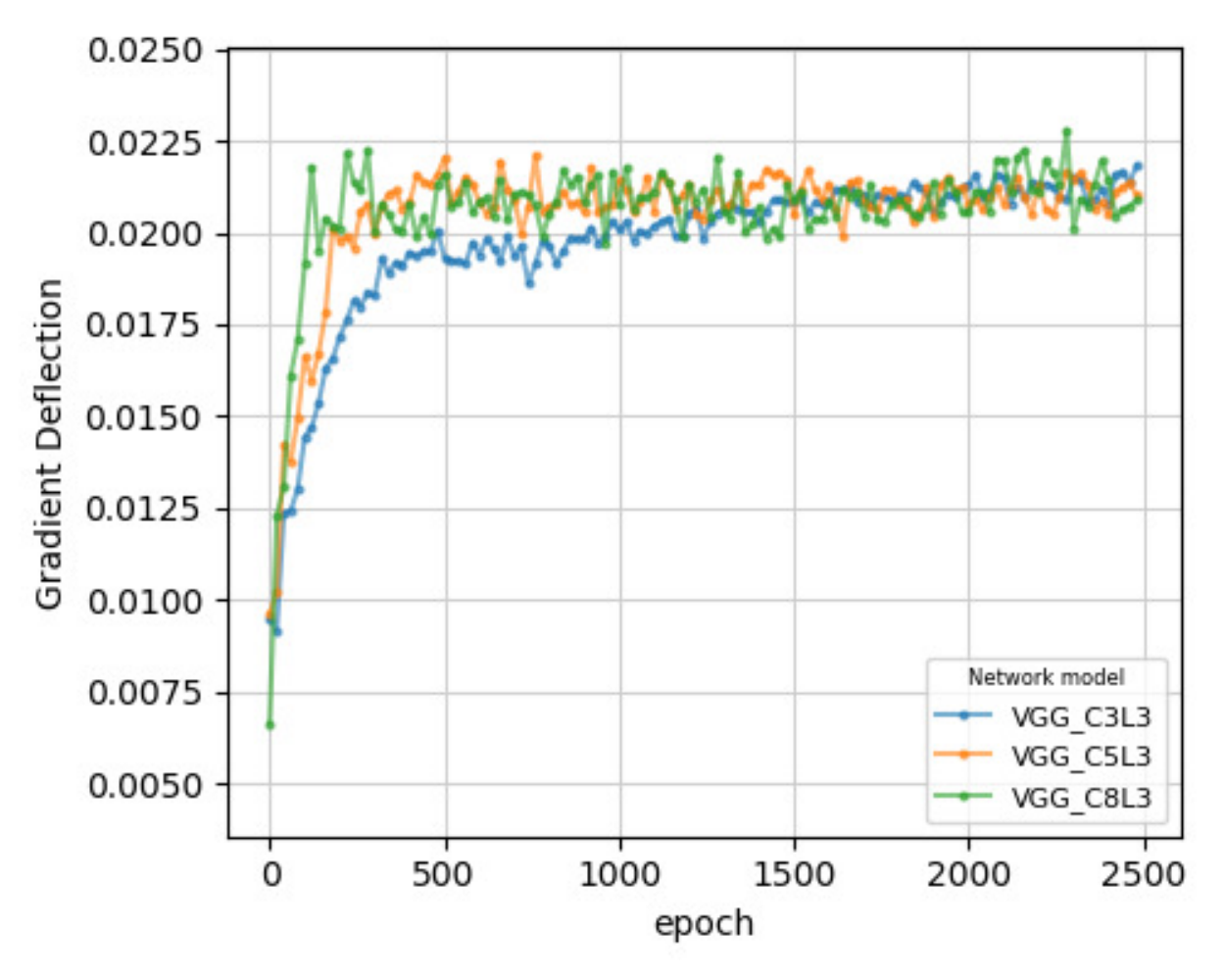}
     \hspace{1.0cm} (f) Gradient Deflection
    \end{center}
   \end{minipage}

  \end{tabular}
  \caption{VGG networks with different number of layers.
  (a-c) Each VGG network in the initial state on linearly interpolating
  paths between each pair of samples in MNIST, CIFAR10 and CIFAR100.
  Plots are averaged over 100 networks
  for each of the different number of layers.
  (d-f) Each VGG network changes with SGD training steps on CIFAR10.}
  \label{fig:VGG}
 \end{center}
\end{figure*}
%
%
%

We also define a NN gradient at $t \in I_k$
as $R_k = \nabla_v u (t)$ for VGG or ResNet.
For each node $t_k$ let $Y_k = R_k - R_{k-1}$,~~$k \in [\mathcal{K}]$
be a gap of NN gradients.
If each $Y_k$ is an i.i.d.\ random variable, then
$R_k, k \in [\mathcal{K}]$ is a random walk bridge
between $R_0, R_{\mathcal{K}}$.
Let $P$ be the distribution of $Y_k$
as the inputs $X_0, X_1~~(X_0 \neq X_1)$
are chosen randomly from training or test samples,
and let $S_k$ be a random walk bridge \eqref{bridge}
generated from i.i.d.\ random variables $Z_k$
with the distribution $P$.
%
As with an MLP, the initial weights determine the gap distribution $P$,
and weights change little in the over-parameterized setting.
Accordingly, $P$ also changes little.

\begin{figure*}[t]
 \begin{center}
  \begin{tabular}{c}
  
   \begin{minipage}{0.32\hsize}
    \begin{center}
     \includegraphics[width=5.0cm]{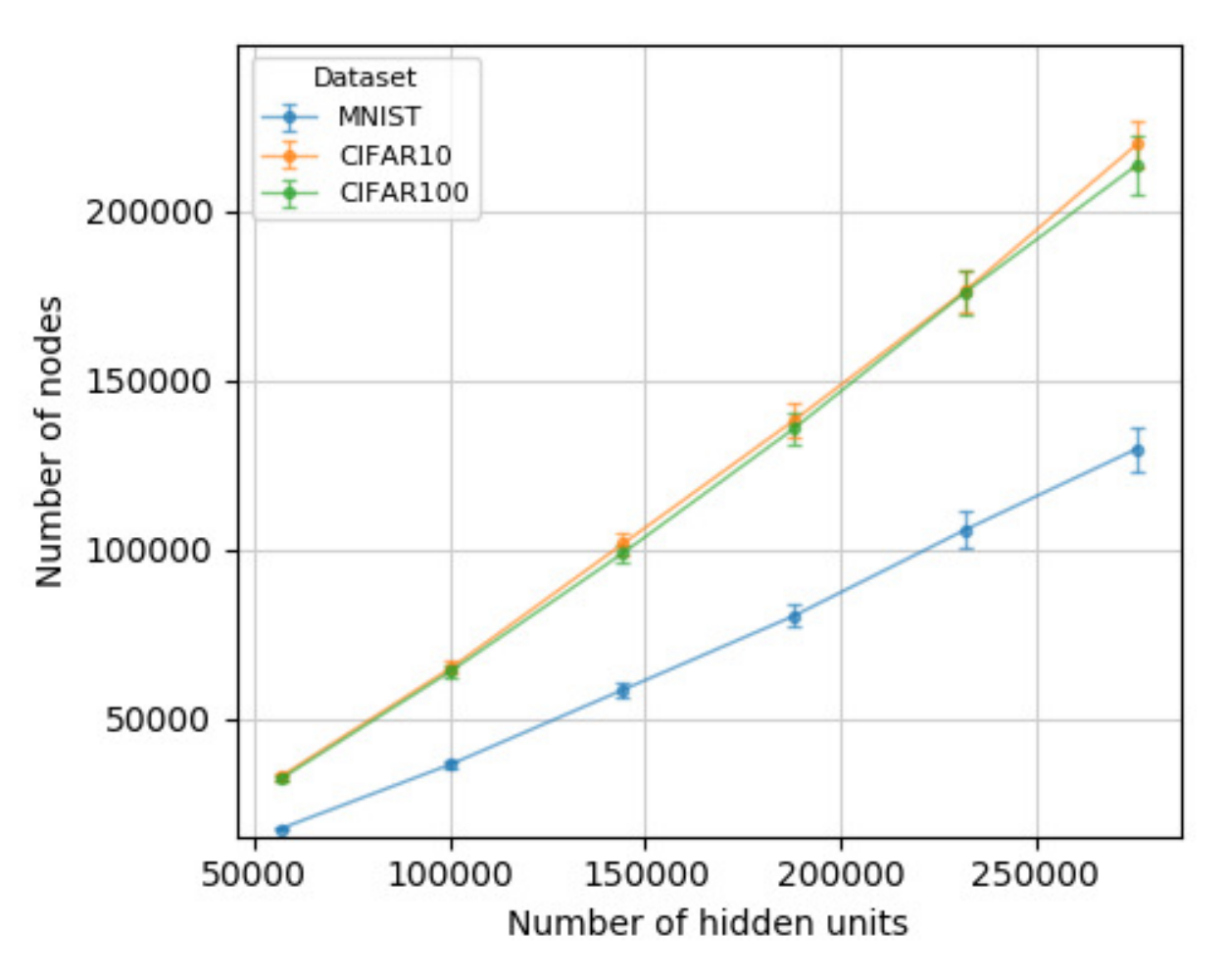}
     \hspace{1.0cm} (a) Number of nodes
    \end{center}
   \end{minipage}

   \begin{minipage}{0.32\hsize}
    \begin{center}
     \includegraphics[width=5.0cm]{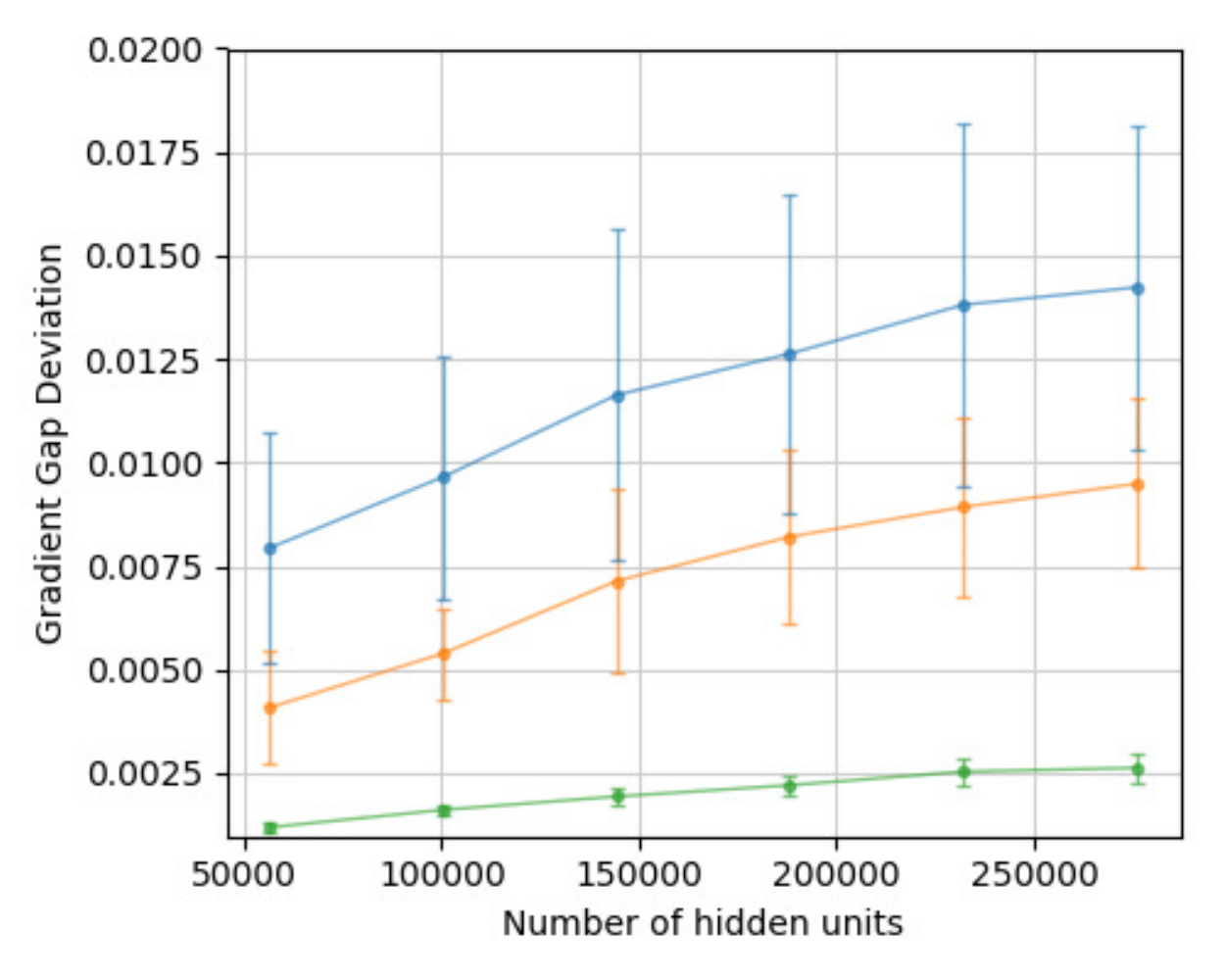}
     \hspace{1.0cm} (b) Gradient Gap Deviation
    \end{center}
   \end{minipage}

   \begin{minipage}{0.32\hsize}
    \begin{center}
     \includegraphics[width=5.0cm]{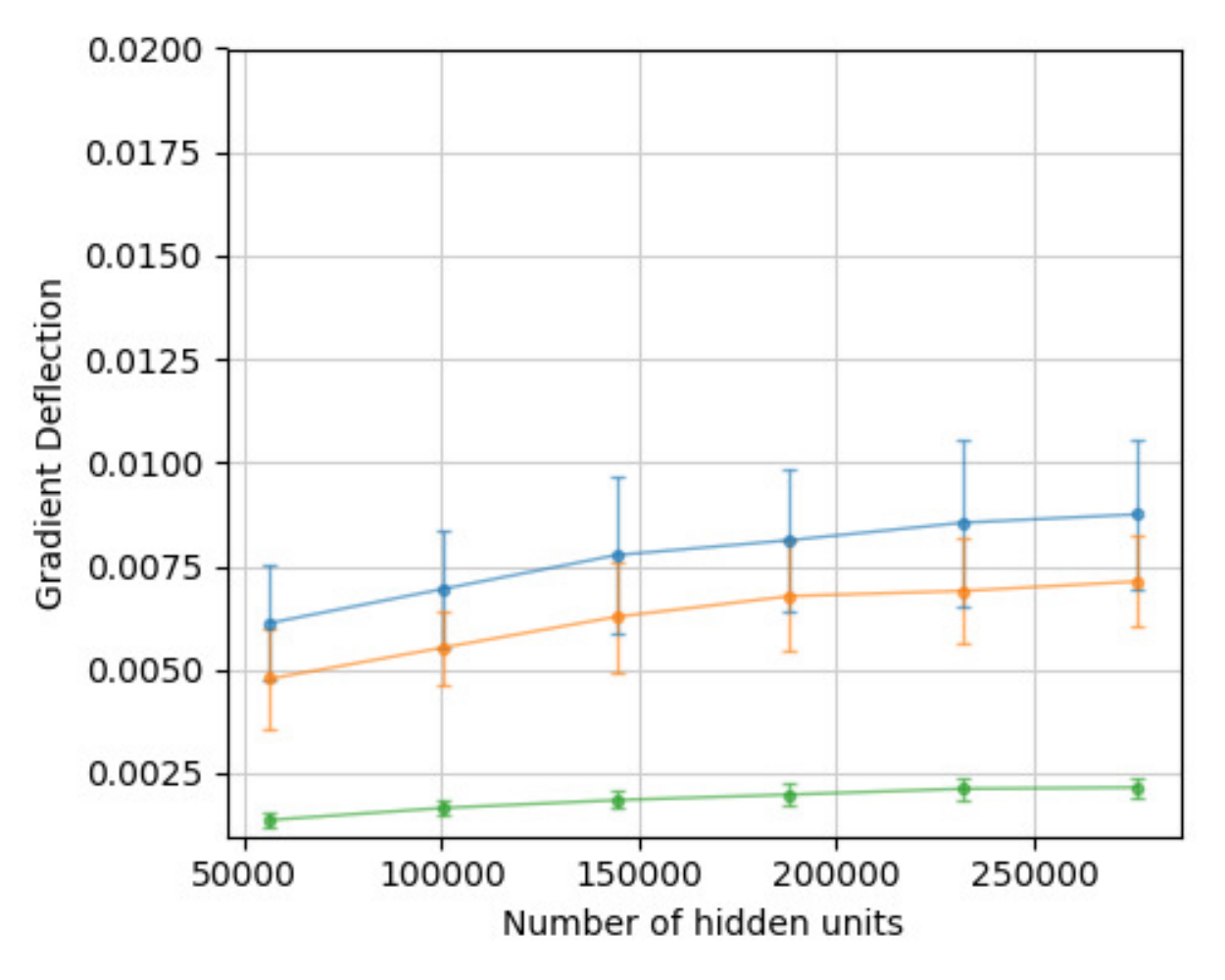}
     \hspace{1.0cm} (c) Gradient Deflection
    \end{center}
   \end{minipage}
   \\~~\\
   \begin{minipage}{0.32\hsize}
    \begin{center}
     \includegraphics[width=5.0cm]{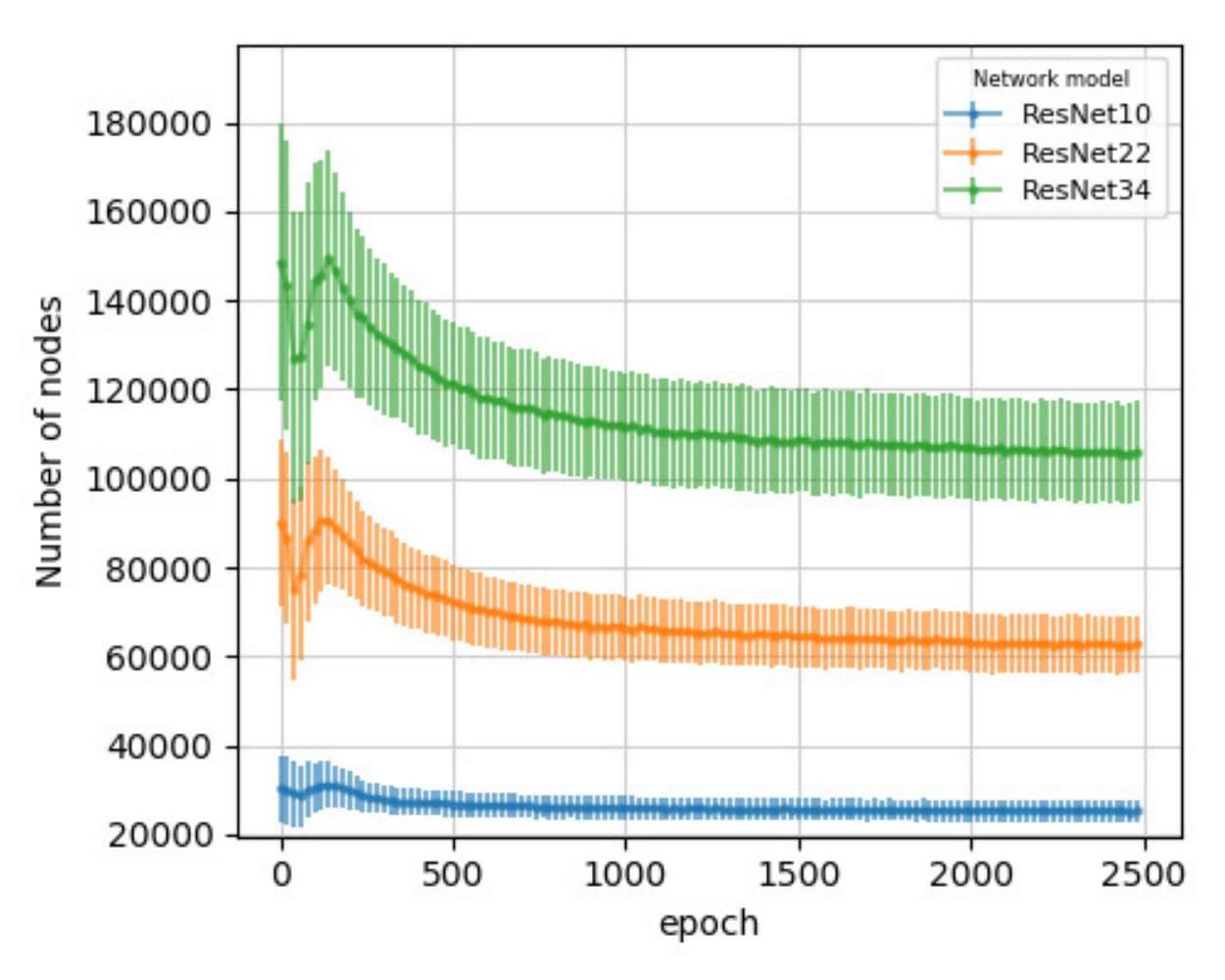}
     \hspace{1.0cm} (d) Number of nodes
    \end{center}
   \end{minipage}

   \begin{minipage}{0.32\hsize}
    \begin{center}
     \includegraphics[width=5.0cm]{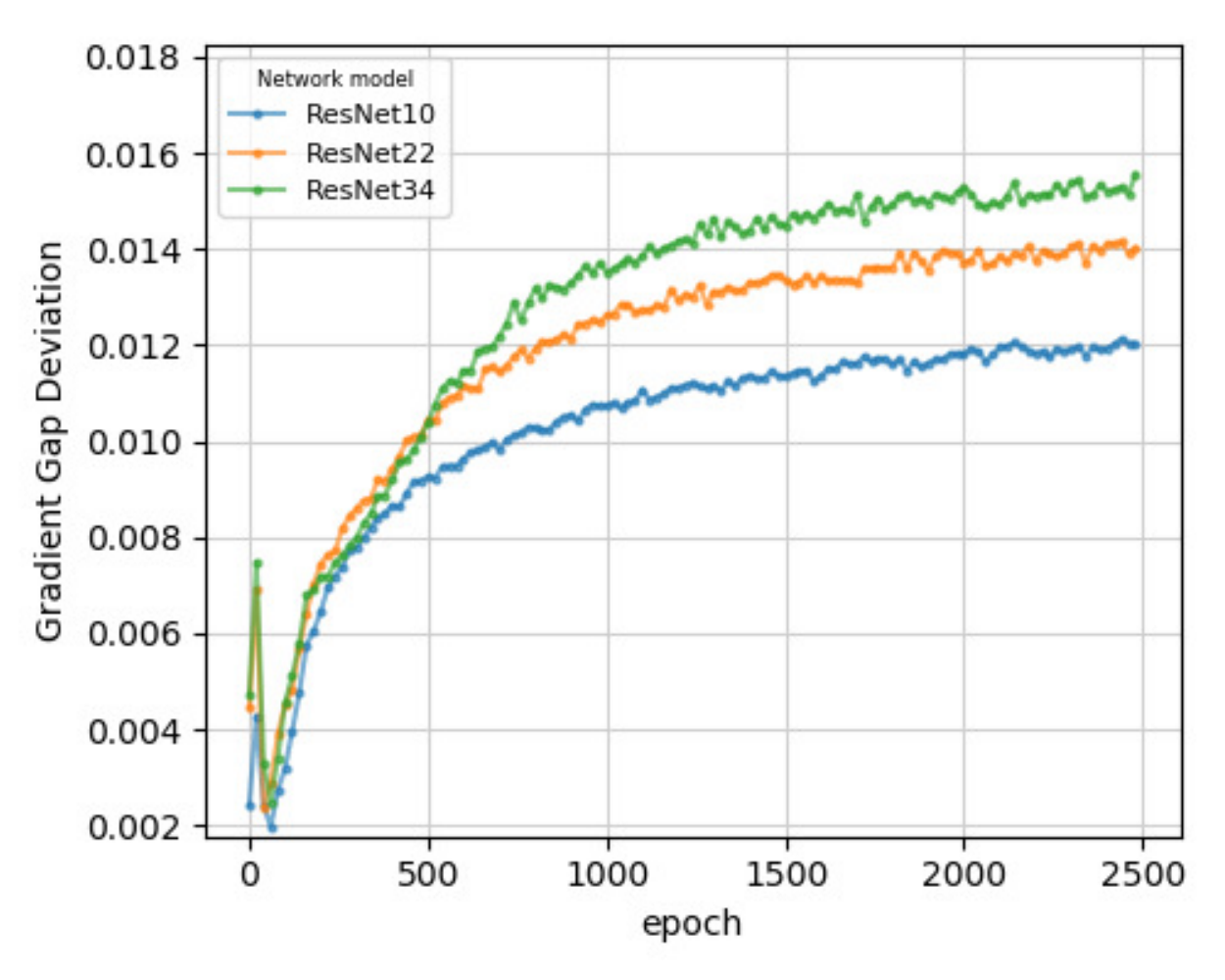}
     \hspace{1.0cm} (e) Gradient Gap Deviation
    \end{center}
   \end{minipage}

   \begin{minipage}{0.32\hsize}
    \begin{center}
     \includegraphics[width=5.0cm]{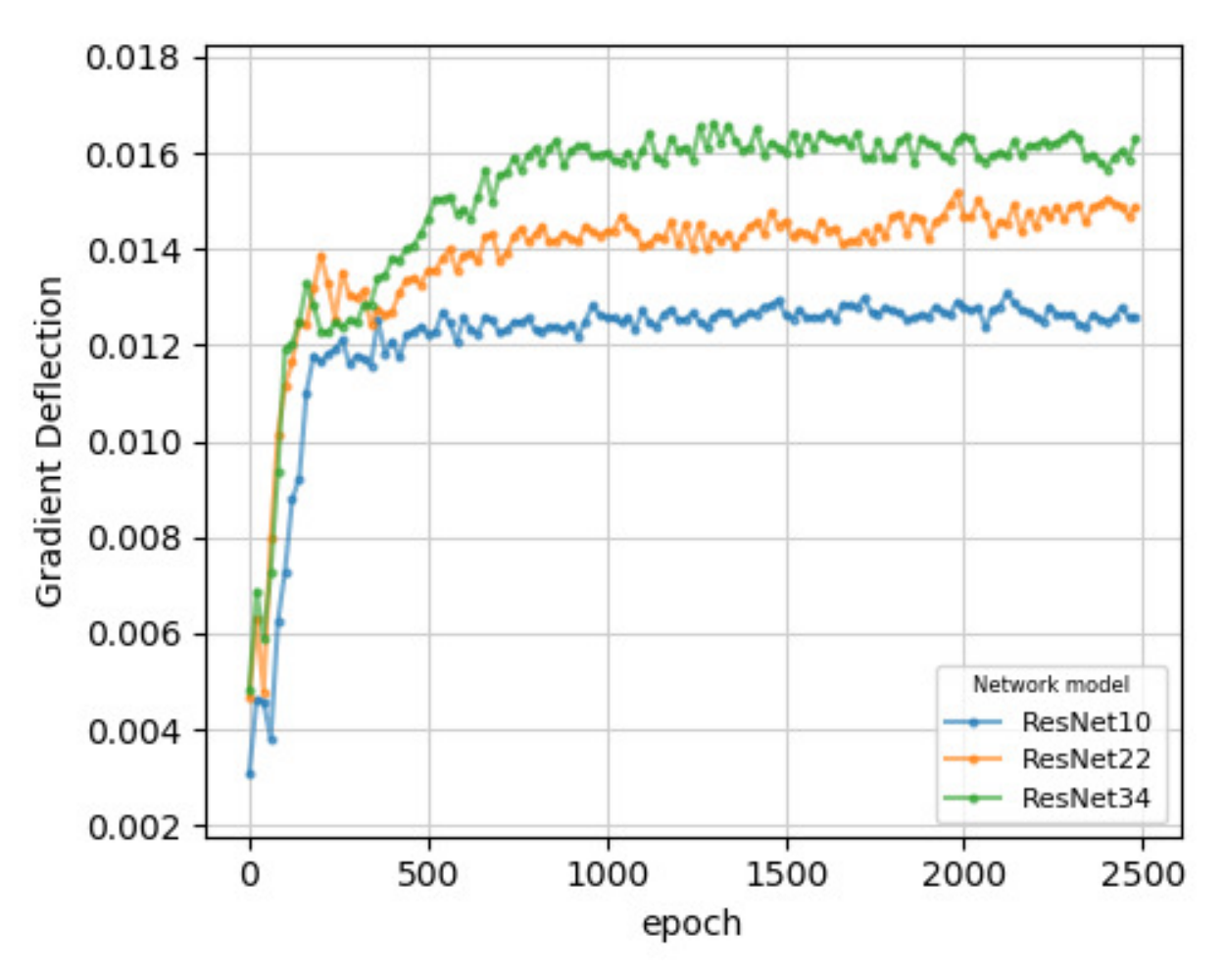}
     \hspace{1.0cm} (f) Gradient Deflection
    \end{center}
   \end{minipage}

  \end{tabular}
  \caption{Residual networks with different number of layers.
  (a-c) Each residual network in the initial state on linearly interpolating
  paths between each pair of samples in MNIST, CIFAR10 and CIFAR100.
  Plots are averaged over 100 networks for each of the
  different number of layers.
  (d-f) Each residual network changes with SGD training steps on CIFAR100.}
  \label{fig:resnet}
 \end{center}
\end{figure*}

First we show experimental results for VGG and ResNet in the initial state
(without training) on MNIST, CIFAR10 and CIFAR100.
Figure \ref{fig:VGG} (a) and Figure \ref{fig:resnet} (a) show
that the number of nodes is only half of
the number of entire hidden units over VGG and ResNet or less.
Figure \ref{fig:VGG} (b-c) for VGG and Figure \ref{fig:resnet} (b-c) for ResNet
show that the values of gradient gap deviation and
gradient deflection are small ($\ll 1.0$).
%
Next, we make use of the normalized NN output $\tilde{f}(W,x)$
to handle the weight change with training.
Figure \ref{fig:VGG} (d) and Figure \ref{fig:resnet} (d) show
the changes in the number of nodes between samples with SGD training steps.
%
%
In order to calculate precise NN gradient and
avoid \textit{cancellation of significant digits},
we also use the method of replacing each ReLU activation layer with
indicator matrix \eqref{indicator}.
%
In Figure \ref{fig:VGG} (e-f) and Figure \ref{fig:resnet} (e-f),
we show the changes in the gradient gap deviation and the gradient deflection
at midpoint $k = \left\lfloor \frac{\mathcal{K}}{2} \right\rfloor$
with SGD training steps.

%
%
%
%
%
%

By measuring the difference between
\textbf{gradient gap deviation} and \textbf{gradient deflection},
it is possible to estimate systematically the variations in output
between samples in comparison with completely random state.
%
In other words, the above difference indicates the degree of
randomness of NN gradients.
%
%
Figure \ref{fig:VGG} (e-f) and Figure \ref{fig:resnet} (e-f) show that
NNs tend to be random between a pair of samples after enough training,
which suggests that NNs become passive or do not control
the network output between samples actively.
%

No matter how large the complexity and expressivity of deep models are,
random variation of the models between samples keeps small $(\ll 1.0)$,
which results in smoothness between a pair of samples
(Figure \ref{fig:VGG} (e-f) and Figure \ref{fig:resnet} (e-f)).
%
This suggests that more expressive power
of deep models than needed is restricted,
which is thought to be one of revealed abilities of
implicit regularization in the over-parameterized regime.
%
We also compare the results on training data with the ones on test data
in Appendix, which suggests that there is no large
difference between training data and test data.

%
%
%
\begin{figure*}[t]
 \begin{center}
  \begin{tabular}{c}
  
   \begin{minipage}{0.32\hsize}
    \begin{center}
     \includegraphics[width=5.0cm]{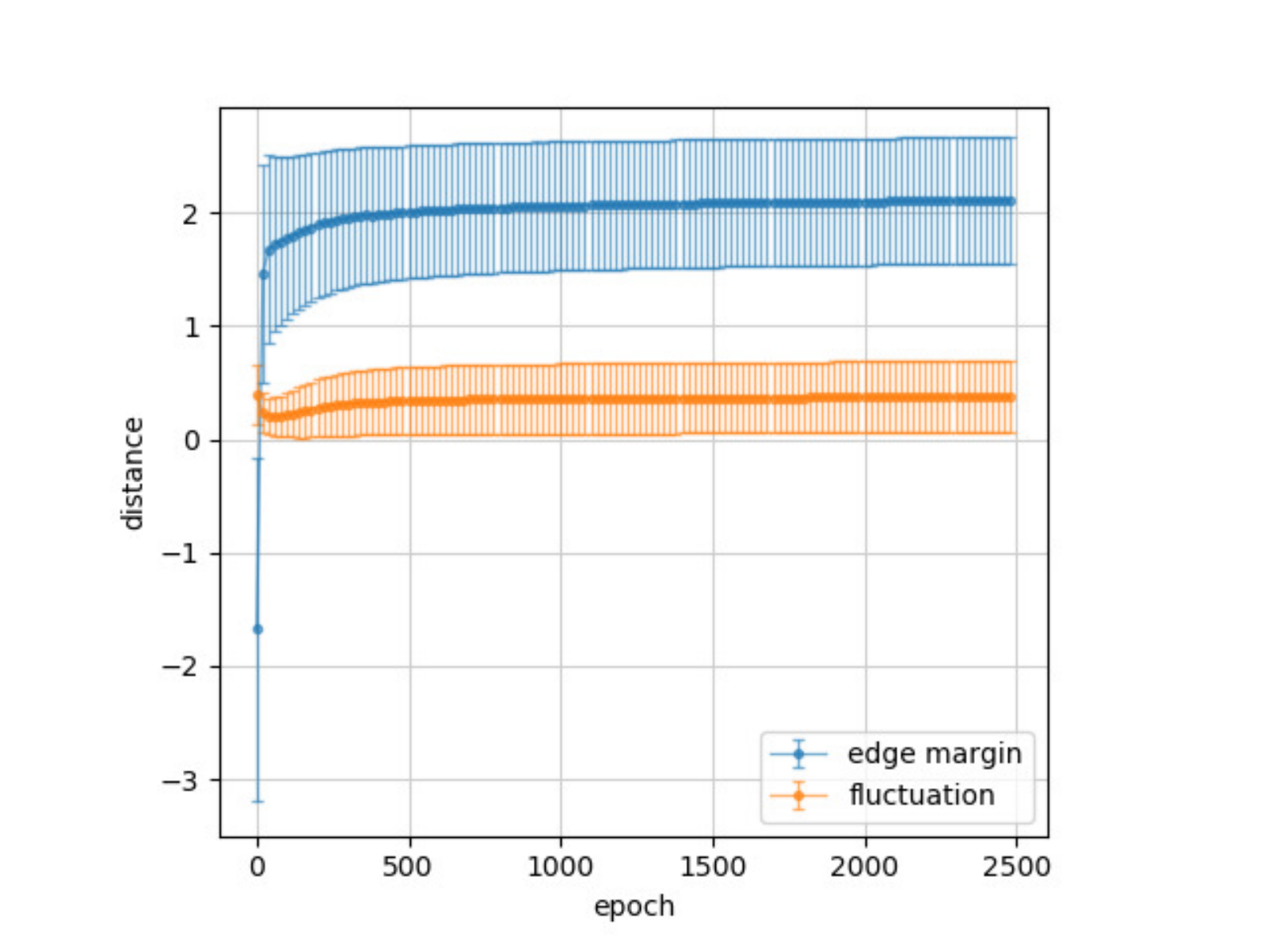}
     \hspace{0.2cm} (a) The 2-layer MLP on MINST
    \end{center}
   \end{minipage}

   \begin{minipage}{0.32\hsize}
    \begin{center}
     \includegraphics[width=5.0cm]{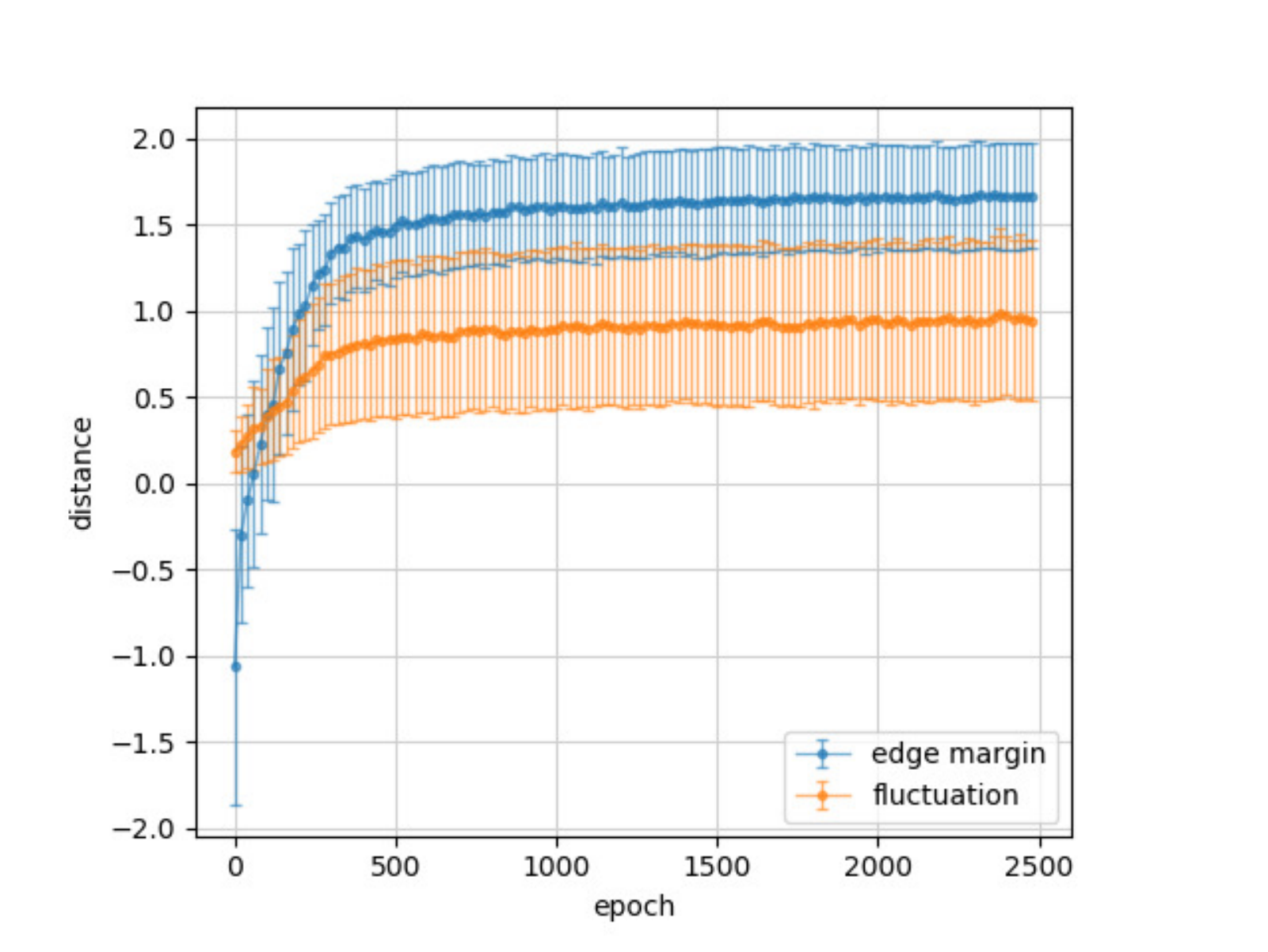}
     \hspace{0.2cm} (b) VGG C8L3 on CIFAR10
    \end{center}
   \end{minipage}

   \begin{minipage}{0.32\hsize}
    \begin{center}
     \includegraphics[width=5.0cm]{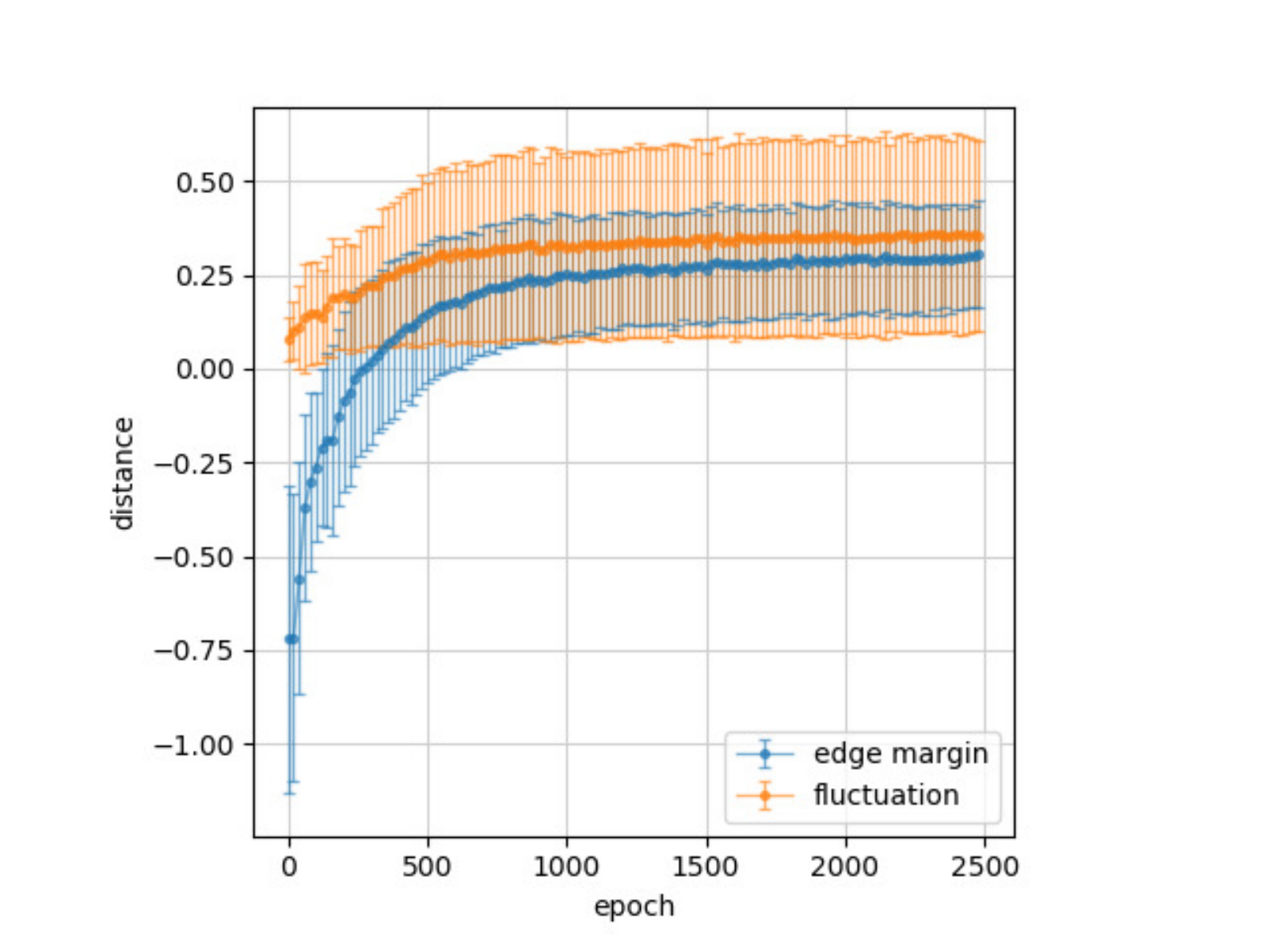}
     \hspace{0.2cm} (c) ResNet34 on CIFAR100
    \end{center}
   \end{minipage}

  \end{tabular}
  \caption{Comparison between pair margin and pair fluctuation during training.
  (a) The 2-layer MLP on MNIST. (b) VGG C8L3
  (8 convolution layer + 3 linear layer) on CIFAR10. 
  (c) ResNet34 on CIFAR100.}
  \label{fig:margin}
 \end{center}
\end{figure*}
%
%
%

In order to estimate the generalization ability
on a linear interpolation between each pair of training samples,
we introduce the following $\boldsymbol{(x_i,x_j)}$
\textit{\textbf{pair margin}} $\boldsymbol{PM_{ij}}$:
%
For the input sample $x_i$, its correct label $y_i$
and the normalized output vector $\tilde{f}(W,x_i)$,
we define $M_i$ as 
$M_i := \tilde{f_{y_i}}(W,x_i) - \max_{j\neq y_i}\tilde{f_j}(W,x_i)$.
%
This value $M_i$ is the margin \cite{bartlett2017spectrally},
which measures the gap between the output for
the correct label and other labels.
%
Then let $PM_{ij}$ be an average of these two values:
$PM_{ij} := (M_i + M_j)/2$.
%
This is corresponding to the value that interpolates
the margin linearly at the midpoint
of $x_i$ and $x_j$ (i.e., $(x_i + x_j)/2$).

Next, we introduce the following $\boldsymbol{(x_i,x_j)}$
\textit{\textbf{pair fluctuation}} $\boldsymbol{PF_{ij}}$
as the amount of fluctuation of the NN output between two samples.
%
Let us denote by $X_0 = x_i$, $X_1 = x_j$ a pair of samples,
and by $u_{ij}(t) := u(t; X_0, X_1)$
the normalized NN output $\tilde{f}(W,X(t))$ between two points $X_0, X_1$.
%
%
We define the pair fluctuation $PF_{ij}$ between the two points $X_0, X_1$
as the difference between the mid point value $u_{ij}(1/2)$ and 
the mean of edge values $u_{ij}(0)$ and $u_{ij}(1)$:
$PF_{ij} := \left| (u_{ij}(0) + u_{ij}(1))/2 - u_{ij}(1/2) \right|$.

Figure \ref{fig:margin} shows the changes in the pair margin $PM$ and
the pair fluctuation $PF$ with SGD training steps.
As the training progresses, the value of $PM$ exceeds
the one of $PF$, which suggests that the accuracy increases
even at the midpoint of two training samples,
and the relation between $PM$ and $PF$ is likely to
be coupled with the validation accuracy.
%


\section{Related Works}

Recently, many works try to explain generalization of neural networks
linked to random initialization and over-parameterization.
%
%
Recent work has shown that
the difference between the learned weights and the initialization
is small compared to the initialization in the over-parameterized regime
\cite{xie2016diverse,tsuchida2017invariance,li2018learning},
and SGD can find global minima of the training objective of
DNNs with ReLU activation in polynomial time
\cite{du2018gradient,allen2018convergencet}.
Our method was inspired by a line of work 
\cite{allen2018convergencer,allen2018convergencet},
which presents a method of estimating weight variation and NN norm.
The work in \cite{allen2018learning} explains implicit regularization
for two-layer NNs without weight regularizer.
%
%
%
Many recent works also study the importance of random initialization
\cite{glorot2010understanding,he2015delving,martens2010deep,sutskever2013importance,sussillo2014random,mishkin2015all,giryes2016deep}.

It was pointed out empirically that
the generalization ability of over-parameterized NNs
cannot be estimated properly in classic learning theory
\cite{neyshabur2014search,zhang2016understanding},
which triggered various research on generalization.
The relationship between loss landscape and generalization
\cite{keskar2016large,dinh2017sharp,wu2017towards,chaudhari2018stochastic},
and the degree of memorization and generalization 
\cite{arpit2017closer}
have revealed.
Several different measures for the generalization capabilities of DNNs
have been examined, and generalization bounds for NNs with ReLU activation
have been presented in terms of the product of the spectral norm and the
Frobenius norm of their weights
\cite{hardt2015train,neyshabur2017pac,bartlett2017spectrally,arora2018stronger}.
Theoretical analysis on generalization on over-parameterized NNs has been
presented
\cite{poggio2017theory,poggio2018theory}.

The work in \cite{sokolic2017robust} shows that bounding
the Frobenius norm of the Jacobian matrix, which is first-order derivatives,
reduces the obtained generalization error.
The work in \cite{novak2018sensitivity} shows that NNs implement
more robust functions in the vicinity of
the training data manifold than away from it,
as measured by the norm of the Jacobian matrix.
While they use the norm of the Jacobian matrix
to estimate generalization and sensitivity,
we evaluate the gradient gap deviation and the gradient deflection
corresponding to Hessian and curvature
to see NN ability in a different light.
We therefore find that although the NN derivative changes
rather randomly between samples, it interpolates
between samples linearly because the amount of its variation is small.

Some recent studies \cite{ma2017power,maennel2018gradient,belkin2018does}
indicate that the generalization error of a NN is small
because a NN interpolates smoothly between a pair of samples.
The work in
\cite{brutzkus2017sgd,soudry2018implicit,li2018learning,perez2018deep}
uses low complexity of the solution found by SGD
to explain the small generalization error of over-parameterized
models measured by classification margin.
However, without any additional assumption on
the dataset structure and the network architecture, 
the direct reason why SGD can find smooth and 
low-complexity solutions is unknown.
In the present work, we tackle this question, and show that
in the general setting, SGD restricts gradient gap and the number of nodes
to be small, and the NN output between samples is properly controlled by
weight initialization and SGD to keep connecting samples almost straight.

The work in \cite{balduzzi2017shattered}
demonstrates the connection between the gradients of a two-layer
fully-connected NN and discrete Brownian motion (random walk).
In this work, we model the NN gradient as a \textbf{random walk bridge}
to estimate the amount of change in the network gradient,
and explain the low complexity of the learned solution,
as measured by gradient gap deviation and gradient deflection.
The work in \cite{raghu2016expressive,novak2018sensitivity} investigates
trajectory length of a one-dimensional path, and finds that
the trajectory length grows exponentially in the depth of the NN.
In contrast, we stochastically estimate gradient variation
between samples and show that the motion of the network gradient
is probably close to a random walk bridge.


\section{Discussion}

We model the gradient of NN output between a pair of samples
as a \textbf{random walk bridge},
and introduce the \textbf{gradient gap deviation} and
the \textbf{gradient deflection}.
%
%
Furthermore, we estimate the global variation of ReLU NNs between samples,
and reveal that ReLU NNs interpolate almost linearly between samples
even in the over-parameterized regime.
%
%
For standard ReLU NNs (MLP, VGG and ResNet) and
datasets (MNIST, CIFAR10 and CIFAR100),
we measure these values and investigate how implicit regularization works.
%
The experimental observations suggest that
the gradient gap deviation and the gradient deflection
are both small for at least the above networks,
which means that the network output interpolates
between samples almost linearly.
%
NNs converge to small fluctuation depending on the dataset,
because excessive complexity and a large degree of freedom
due to over-parameterization are controlled by SGD properly
(we also ensured this result is independent of whether
we use Momentum SGD or vanilla SGD).
%
In order to keep the NN fluctuation to be small,
it is necessary to prevent the gradient gap and the number of nodes
from being larger than needed.
%
In other words, one of the mechanisms of implicit regularization by SGD
seems to restrict the gradient gap and the number of nodes.
%

One of the interesting questions for future work is to prove that
for more deep networks,
the number of nodes is only half of
the number of entire hidden units or less.
Our work is a step towards theoretical understanding of
implicit regularization by random initialization and 
optimization algorithm for neural networks,
and the study of constitutive relations between
generalization and implicit regularization is left for future work.


\bibliographystyle{plain}

\bibliography{A2_arXiv}

\begin{thebibliography}{10}

\bibitem{allen2018learning}
Zeyuan Allen-Zhu, Yuanzhi Li, and Yingyu Liang.
\newblock Learning and generalization in overparameterized neural networks,
  going beyond two layers.
\newblock {\em arXiv preprint arXiv:1811.04918}, 2018.

\bibitem{allen2018convergencet}
Zeyuan Allen-Zhu, Yuanzhi Li, and Zhao Song.
\newblock A convergence theory for deep learning via over-parameterization.
\newblock {\em arXiv preprint arXiv:1811.03962}, 2018.

\bibitem{allen2018convergencer}
Zeyuan Allen-Zhu, Yuanzhi Li, and Zhao Song.
\newblock On the convergence rate of training recurrent neural networks.
\newblock {\em arXiv preprint arXiv:1810.12065}, 2018.

\bibitem{arora2018stronger}
Sanjeev Arora, Rong Ge, Behnam Neyshabur, and Yi~Zhang.
\newblock Stronger generalization bounds for deep nets via a compression
  approach.
\newblock {\em arXiv preprint arXiv:1802.05296}, 2018.

\bibitem{arpit2017closer}
Devansh Arpit, Stanis{\l{}}aw Jastrz{\k{e}}bski, Nicolas Ballas, David Krueger,
  Emmanuel Bengio, Maxinder~S Kanwal, Tegan Maharaj, Asja Fischer, Aaron
  Courville, Yoshua Bengio, et~al.
\newblock A closer look at memorization in deep networks.
\newblock {\em arXiv preprint arXiv:1706.05394}, 2017.

\bibitem{balduzzi2017shattered}
David Balduzzi, Marcus Frean, Lennox Leary, JP~Lewis, Kurt Wan-Duo Ma, and
  Brian McWilliams.
\newblock The shattered gradients problem: If resnets are the answer, then what
  is the question?
\newblock {\em arXiv preprint arXiv:1702.08591}, 2017.

\bibitem{bartlett2017spectrally}
Peter~L Bartlett, Dylan~J Foster, and Matus~J Telgarsky.
\newblock Spectrally-normalized margin bounds for neural networks.
\newblock In {\em Advances in Neural Information Processing Systems}, pages
  6240--6249, 2017.

\bibitem{belkin2018does}
Mikhail Belkin, Alexander Rakhlin, and Alexandre~B Tsybakov.
\newblock Does data interpolation contradict statistical optimality?
\newblock {\em arXiv preprint arXiv:1806.09471}, 2018.

\bibitem{brutzkus2017sgd}
Alon Brutzkus, Amir Globerson, Eran Malach, and Shai Shalev-Shwartz.
\newblock Sgd learns over-parameterized networks that provably generalize on
  linearly separable data.
\newblock {\em arXiv preprint arXiv:1710.10174}, 2017.

\bibitem{chaudhari2018stochastic}
Pratik Chaudhari and Stefano Soatto.
\newblock Stochastic gradient descent performs variational inference, converges
  to limit cycles for deep networks.
\newblock In {\em 2018 Information Theory and Applications Workshop (ITA)},
  pages 1--10. IEEE, 2018.

\bibitem{cohen2016expressive}
Nadav Cohen, Or~Sharir, and Amnon Shashua.
\newblock On the expressive power of deep learning: A tensor analysis.
\newblock In {\em Conference on Learning Theory}, pages 698--728, 2016.

\bibitem{dinh2017sharp}
Laurent Dinh, Razvan Pascanu, Samy Bengio, and Yoshua Bengio.
\newblock Sharp minima can generalize for deep nets.
\newblock {\em arXiv preprint arXiv:1703.04933}, 2017.

\bibitem{du2018gradient}
Simon~S Du, Xiyu Zhai, Barnabas Poczos, and Aarti Singh.
\newblock Gradient descent provably optimizes over-parameterized neural
  networks.
\newblock {\em arXiv preprint arXiv:1810.02054}, 2018.

\bibitem{giryes2016deep}
Raja Giryes, Guillermo Sapiro, and Alexander~M Bronstein.
\newblock Deep neural networks with random gaussian weights: A universal
  classification strategy?
\newblock {\em IEEE Trans. Signal Processing}, 64(13):3444--3457, 2016.

\bibitem{glorot2010understanding}
Xavier Glorot and Yoshua Bengio.
\newblock Understanding the difficulty of training deep feedforward neural
  networks.
\newblock In {\em Proceedings of the thirteenth international conference on
  artificial intelligence and statistics}, pages 249--256, 2010.

\bibitem{hardt2015train}
Moritz Hardt, Benjamin Recht, and Yoram Singer.
\newblock Train faster, generalize better: Stability of stochastic gradient
  descent.
\newblock {\em arXiv preprint arXiv:1509.01240}, 2015.

\bibitem{he2015delving}
Kaiming He, Xiangyu Zhang, Shaoqing Ren, and Jian Sun.
\newblock Delving deep into rectifiers: Surpassing human-level performance on
  imagenet classification.
\newblock In {\em Proceedings of the IEEE international conference on computer
  vision}, pages 1026--1034, 2015.

\bibitem{he2016deep}
Kaiming He, Xiangyu Zhang, Shaoqing Ren, and Jian Sun.
\newblock Deep residual learning for image recognition.
\newblock In {\em Proceedings of the IEEE conference on computer vision and
  pattern recognition}, pages 770--778, 2016.

\bibitem{he2016identity}
Kaiming He, Xiangyu Zhang, Shaoqing Ren, and Jian Sun.
\newblock Identity mappings in deep residual networks.
\newblock In {\em European Conference on Computer Vision}, pages 630--645.
  Springer, 2016.

\bibitem{keskar2016large}
Nitish~Shirish Keskar, Dheevatsa Mudigere, Jorge Nocedal, Mikhail Smelyanskiy,
  and Ping Tak~Peter Tang.
\newblock On large-batch training for deep learning: Generalization gap and
  sharp minima.
\newblock {\em arXiv preprint arXiv:1609.04836}, 2016.

\bibitem{krizhevsky2009learning}
Alex Krizhevsky and Geoffrey Hinton.
\newblock Learning multiple layers of features from tiny images.
\newblock Technical report, Citeseer, 2009.

\bibitem{lecun1998gradient}
Yann LeCun, L{\'e}on Bottou, Yoshua Bengio, Patrick Haffner, et~al.
\newblock Gradient-based learning applied to document recognition.
\newblock {\em Proceedings of the IEEE}, 86(11):2278--2324, 1998.

\bibitem{lee2017deep}
Jaehoon Lee, Yasaman Bahri, Roman Novak, Samuel~S Schoenholz, Jeffrey
  Pennington, and Jascha Sohl-Dickstein.
\newblock Deep neural networks as gaussian processes.
\newblock {\em arXiv preprint arXiv:1711.00165}, 2017.

\bibitem{li2018learning}
Yuanzhi Li and Yingyu Liang.
\newblock Learning overparameterized neural networks via stochastic gradient
  descent on structured data.
\newblock In {\em Advances in Neural Information Processing Systems}, pages
  8167--8176, 2018.

\bibitem{liao2018surprising}
Qianli Liao, Brando Miranda, Andrzej Banburski, Jack Hidary, and Tomaso Poggio.
\newblock A surprising linear relationship predicts test performance in deep
  networks.
\newblock {\em arXiv preprint arXiv:1807.09659}, 2018.

\bibitem{liggett1968invariance}
Thomas~M Liggett.
\newblock An invariance principle for conditioned sums of independent random
  variables.
\newblock {\em Journal of Mathematics and Mechanics}, 18(6):559--570, 1968.

\bibitem{ma2017power}
Siyuan Ma, Raef Bassily, and Mikhail Belkin.
\newblock The power of interpolation: Understanding the effectiveness of sgd in
  modern over-parametrized learning.
\newblock {\em arXiv preprint arXiv:1712.06559}, 2017.

\bibitem{maennel2018gradient}
Hartmut Maennel, Olivier Bousquet, and Sylvain Gelly.
\newblock Gradient descent quantizes relu network features.
\newblock {\em arXiv preprint arXiv:1803.08367}, 2018.

\bibitem{martens2010deep}
James Martens.
\newblock Deep learning via hessian-free optimization.
\newblock In {\em ICML}, volume~27, pages 735--742, 2010.

\bibitem{matthews2018gaussian}
Alexander G de~G Matthews, Mark Rowland, Jiri Hron, Richard~E Turner, and
  Zoubin Ghahramani.
\newblock Gaussian process behaviour in wide deep neural networks.
\newblock {\em arXiv preprint arXiv:1804.11271}, 2018.

\bibitem{mishkin2015all}
Dmytro Mishkin and Jiri Matas.
\newblock All you need is a good init.
\newblock {\em arXiv preprint arXiv:1511.06422}, 2015.

\bibitem{montufar2014number}
Guido~F Montufar, Razvan Pascanu, Kyunghyun Cho, and Yoshua Bengio.
\newblock On the number of linear regions of deep neural networks.
\newblock In {\em Advances in neural information processing systems}, pages
  2924--2932, 2014.

\bibitem{nagarajan2017generalization}
Vaishnavh Nagarajan and J~Zico Kolter.
\newblock Generalization in deep networks: The role of distance from
  initialization.
\newblock In {\em NIPS workshop on Deep Learning: Bridging Theory and
  Practice}, 2017.

\bibitem{Neal1994phdthesis}
Radford~M. Neal.
\newblock {\em Bayesian Learning for Neural Networks}.
\newblock PhD thesis, University of Toronto, Dept. of Computer Science, 1994.

\bibitem{neyshabur2017pac}
Behnam Neyshabur, Srinadh Bhojanapalli, David McAllester, and Nathan Srebro.
\newblock A pac-bayesian approach to spectrally-normalized margin bounds for
  neural networks.
\newblock {\em arXiv preprint arXiv:1707.09564}, 2017.

\bibitem{neyshabur2014search}
Behnam Neyshabur, Ryota Tomioka, and Nathan Srebro.
\newblock In search of the real inductive bias: On the role of implicit
  regularization in deep learning.
\newblock {\em arXiv preprint arXiv:1412.6614}, 2014.

\bibitem{novak2018sensitivity}
Roman Novak, Yasaman Bahri, Daniel~A Abolafia, Jeffrey Pennington, and Jascha
  Sohl-Dickstein.
\newblock Sensitivity and generalization in neural networks: an empirical
  study.
\newblock {\em arXiv preprint arXiv:1802.08760}, 2018.

\bibitem{perez2018deep}
Guillermo~Valle P{\'e}rez, Ard~A Louis, and Chico~Q Camargo.
\newblock Deep learning generalizes because the parameter-function map is
  biased towards simple functions.
\newblock {\em arXiv preprint arXiv:1805.08522}, 2018.

\bibitem{poggio2017theory}
Tomaso Poggio, Kenji Kawaguchi, Qianli Liao, Brando Miranda, Lorenzo Rosasco,
  Xavier Boix, Jack Hidary, and Hrushikesh Mhaskar.
\newblock Theory of deep learning iii: explaining the non-overfitting puzzle.
\newblock {\em arXiv preprint arXiv:1801.00173}, 2018.

\bibitem{poggio2018theory}
Tomaso Poggio, Qianli Liao, Brando Miranda, Andrzej Banburski, Xavier Boix, and
  Jack Hidary.
\newblock Theory iiib: Generalization in deep networks.
\newblock {\em arXiv preprint arXiv:1806.11379}, 2018.

\bibitem{poole2016exponential}
Ben Poole, Subhaneil Lahiri, Maithra Raghu, Jascha Sohl-Dickstein, and Surya
  Ganguli.
\newblock Exponential expressivity in deep neural networks through transient
  chaos.
\newblock In {\em Advances in neural information processing systems}, pages
  3360--3368, 2016.

\bibitem{raghu2016expressive}
Maithra Raghu, Ben Poole, Jon Kleinberg, Surya Ganguli, and Jascha
  Sohl-Dickstein.
\newblock On the expressive power of deep neural networks.
\newblock {\em arXiv preprint arXiv:1606.05336}, 2016.

\bibitem{serra2017bounding}
Thiago Serra, Christian Tjandraatmadja, and Srikumar Ramalingam.
\newblock Bounding and counting linear regions of deep neural networks.
\newblock {\em arXiv preprint arXiv:1711.02114}, 2017.

\bibitem{simonyan2014very}
Karen Simonyan and Andrew Zisserman.
\newblock Very deep convolutional networks for large-scale image recognition.
\newblock {\em arXiv preprint arXiv:1409.1556}, 2014.

\bibitem{sokolic2017robust}
Jure Sokoli{\'c}, Raja Giryes, Guillermo Sapiro, and Miguel~RD Rodrigues.
\newblock Robust large margin deep neural networks.
\newblock {\em IEEE Transactions on Signal Processing}, 65(16):4265--4280,
  2017.

\bibitem{soudry2018implicit}
Daniel Soudry, Elad Hoffer, Mor~Shpigel Nacson, Suriya Gunasekar, and Nathan
  Srebro.
\newblock The implicit bias of gradient descent on separable data.
\newblock {\em Journal of Machine Learning Research}, 19(70), 2018.

\bibitem{sussillo2014random}
David Sussillo and LF~Abbott.
\newblock Random walk initialization for training very deep feedforward
  networks.
\newblock {\em arXiv preprint arXiv:1412.6558}, 2014.

\bibitem{sutskever2013importance}
Ilya Sutskever, James Martens, George Dahl, and Geoffrey Hinton.
\newblock On the importance of initialization and momentum in deep learning.
\newblock In {\em International conference on machine learning}, pages
  1139--1147, 2013.

\bibitem{telgarsky2016benefits}
Matus Telgarsky.
\newblock Benefits of depth in neural networks.
\newblock {\em arXiv preprint arXiv:1602.04485}, 2016.

\bibitem{tsuchida2017invariance}
Russell Tsuchida, Farbod Roosta-Khorasani, and Marcus Gallagher.
\newblock Invariance of weight distributions in rectified mlps.
\newblock {\em arXiv preprint arXiv:1711.09090}, 2017.

\bibitem{wu2017towards}
Lei Wu, Zhanxing Zhu, and Weinan E.
\newblock Towards understanding generalization of deep learning: Perspective of
  loss landscapes.
\newblock {\em arXiv preprint arXiv:1706.10239}, 2017.

\bibitem{xiao2018dynamical}
Lechao Xiao, Yasaman Bahri, Jascha Sohl-Dickstein, Samuel~S Schoenholz, and
  Jeffrey Pennington.
\newblock Dynamical isometry and a mean field theory of cnns: How to train
  10,000-layer vanilla convolutional neural networks.
\newblock {\em arXiv preprint arXiv:1806.05393}, 2018.

\bibitem{xie2016diverse}
Bo~Xie, Yingyu Liang, and Le~Song.
\newblock Diverse neural network learns true target functions.
\newblock {\em arXiv preprint arXiv:1611.03131}, 2016.

\bibitem{zhang2016understanding}
Chiyuan Zhang, Samy Bengio, Moritz Hardt, Benjamin Recht, and Oriol Vinyals.
\newblock Understanding deep learning requires rethinking generalization.
\newblock {\em arXiv preprint arXiv:1611.03530}, 2016.

\bibitem{zhang2019fixup}
Hongyi Zhang, Yann Dauphin, and Tengyu Ma.
\newblock Fixup initialization: Residual learning without normalization.
\newblock {\em arXiv preprint arXiv:1901.09321}, 2019.

\end{thebibliography}

\appendix

\section{Proofs}

\subsection{Proof of Theorem 3.1}

By assumption we know that
$X_0$, $X_1$ $\in \R^{d}$ are random vectors with i.i.d.\ entries 
$(X_0)_{i}$, $(X_1)_{i}$ $\sim \mathcal{N}(0,1)$,
and the weight matrix of the first layer $W_1$ is initialized with
$(W_1)_{i,j} \sim \mathcal{N}(0, 2/d)$.\\
Now for fixed $i~~(1 \le i \le m) $, let us denote
the $i^{\text{th}}$ row of $W_1$ as
\[
 W_1^{(i)} = ((W_1)_{i1}, (W_1)_{i2}, \ldots ,(W_1)_{id}) 
\]
We know that with high probability, for large enough input dimension $d>0$,
\[
 2 - \varepsilon \le \|W_1^{(i)} \|_2^2 \le 2 + \varepsilon
\]
We define input vectors of hidden units $h^{(0)}$ and $h^{(1)}$ as follows:
$h^{(0)} := W_1 X_0$, $h^{(1)} := W_1 X_1$.\\
Then we obtain
$h_i^{(\alpha)}$ $\sim \mathcal{N}\left(0, \|W_1^{(i)}\|_2^2 \right)$
$~(\alpha = 0,1)$,
and the two random variables ($h_i^{(0)}$ and $h_i^{(1)}$) are independent.\\
Thus, it indicates that the signs of input $h_i^{(0)}$ and $h_i^{(1)}$ 
are also independent, and each sign is chosen with half probability.\\ 
This means that the number of coincidences of
the two signs, namely, \textsf{\#}$\{i: \sgn(h_i^{(0)}) = \sgn(h_i^{(1)}) \}$,
is in distribution identical to binomial distribution
$\mathcal{B} (m,1/2)$ with $m$ trials and $1/2$ success rate.\\
We complete the proof.

\subsection{Proof of Theorem 3.2}
By assumption we know that $E[ S_k^2 ] = k \sigma^2$.\\
Then we obtain by simple calculation
\begin{align*}
 Var[T_k] = E[T_k^2] 
 &= E[S_k^2] + \frac{k^2}{{\mathcal{K}}^2} E[S_{\mathcal{K}}^2] 
 - \frac{2k}{\mathcal{K}} E[S_k \cdot S_{\mathcal{K}}]\\
 &= E[S_k^2] + \frac{k^2}{{\mathcal{K}}^2} E[S_{\mathcal{K}}^2] 
 - \frac{2k}{\mathcal{K}} E[S_k^2]\\
 & = k\sigma^2 + \frac{k^2}{{\mathcal{K}}^2}\cdot \mathcal{K} \sigma^2 
 - \frac{2k}{\mathcal{K}} \cdot k \sigma^2 \\
 &= k \left( 1- \frac{k}{\mathcal{K}} \right) \sigma^2
\end{align*}
We complete the proof.

\subsection{Proof of Theorem 3.3}

We define a linear interpolation $X(t)$ of two points
$X_0, X_1 \in \R^d  ~~(X_0 \neq X_1)$,
which are chosen from training data or test data.
For a parameter $t \in [0,1]$,
let us denote the interpolation as $X(t) = (1-t) \cdot X_0 + t \cdot X_1$ and
its direction vector as $v = X_1 - X_0$.
We define a vector valued function $U(t) :=  f(W,X(t))$.

Let us assume that $t_* \in [0,1]$ is a \textbf{nodes} of $U(t)$, and 
$G(X(t)) := \{ G_l(X(t))\}_{l=1}^{L-1}$
is the \textbf{indicator matrices} \eqref{indicator} 
defined with respect to $W$.

By the randomness of the weight matrices $W$,
for small enough $\delta >0$, it can be shown that
with probability $1$, the difference between indicator matrices
$G(X(t_* - \delta))$ and $G(X(t_* + \delta))$ is only one element
of one indicator matrix of a certain layer $l \in [L-1]$
(i.e., one element difference between $G_{l}(X(t_* - \delta))$ and
$G_{l}(X(t_* + \delta))$ for some $l \in [L-1]$).

Let us denote $p = t_* - \delta$ and $q = t_* + \delta$.
For simplicity, we assume that the difference between 
indicator matrices $G(X(p))$ and $G(X(q))$
is only one element difference between $G_{l}(X(p))$ and $G_{l}(X(q))$.

For $t = p$, let us denote $G_j := G_j(X(p)) ~~(1\le j \le L)$.
Then we have
\[
 U(p) = W_L G_L  \cdots W_{l+1} G_l W_l \cdots G_1 W_1 X(p).
\]

Under the above assumption, we set ${G_l}':= G_{l}(X(q))$ and obtain 
\[
 U(q) = W_L G_L  \cdots  W_{l+1} {G_l}' W_l \cdots G_1 W_1 X(q).
\]

Since $U(t)$ is linear on a small neighborhood of $p$,
the gradient at $p$ 
$\displaystyle \left( \nabla_v U (p) := \lim_{r \to 0} \frac{U(p + r) - U(p)}{h\|v\|}\right)$
is equal to
$\displaystyle \frac{U(p + \varepsilon) - U(p)}{h\|v\|}$ 
for small enough $\varepsilon > 0$, and
we obtain
\begin{equation}
 \label{product of matrix}
 \nabla_v U (p) = W_L G_L  \cdots W_{l+1} G_l W_l \cdots G_1 W_1 \xi, 
\end{equation}
where $\displaystyle \xi := \frac{v}{\| v \|}$ is the normalized vector.

Similarly, we know that
$\displaystyle \nabla_v U (q) := \frac{U(q + \varepsilon) - U(q)}{h\|v\|}$
is equal to the following:
\[
 \nabla_v U (q) = W_L G_L  \cdots W_{l+1} {G_l}' W_l \cdots G_1 W_1 \xi
\]

This implies that the gradient gap $M$ at $t =t_*$ is as follows:
\begin{align*}
 M :=&~  \nabla_v U (q) - \nabla_v U (p) \\
 ~  =&~ W_L G_L \cdots W_{l+1}({G_l}'- G_l)W_l \cdots G_1 W_1 \xi.
\end{align*}

For small enough $\varepsilon > 0$, 
only one element of the matrix ${G_l}'- G_l$ is non-zero, and the value
of the element is equal to $\pm 1$.

Next, in order to estimate the gradient gap $M$,
using concentration inequalities,
we obtain the following estimate
(see Lemma 4.1 in \cite{allen2018convergencet}):\\
\noindent
\textbf{Lemma}. If $\varepsilon  \in (0,1]$, with probability at least 
$1 - e^{- \Omega(m\varepsilon^2/L)}$ over the randomness $W$,
for a fixed unit vector $z_{a-1}$ and $a,b \in [L]$ with $a < b$,
we have
\[
\|  G_b W_b \cdots  G_a W_a z_{a-1} \|_2^2
~\in~ [1-\varepsilon, 1+ \varepsilon].
\]

This Lemma implies that with probability at least 
$1 - e^{- \Omega(m\varepsilon^2/L)}$ over the randomness $W$,
we have
\[
 \| z_{l-1}\|_2^2 := \| G_{l-1} W_{l-1} \cdots  G_1 W_1 \xi \|_2^2
 ~\in~ [1-\varepsilon, 1+ \varepsilon].
\]

In the following, for a fixed unit vector $z_{l-1}$, 
we will estimate the $L^2$ norm of the $l^{\text{th}}$ layer activation
\[
 z_{l} := ({G_l}'- G_l)W_l z_{l-1}.
\]

(Remark: If $l = 1$, then  $z_{l-1}$ is equal to the unit vector $\xi$.)

By the assumption, except for the one element of the matrix ${G_l}'- G_l$,
all elements of the matrix ${G_l}'- G_l$ are equal to zero. 
Thus, without loss of generality, we assume that the $k^{\text{th}}$ 
diagonal element is not equal to zero.

This implies that the nonzero element of $({G_l}'- G_l)W_l z_{l-1}$
is equal to the $k^{\text{th}}$ component of $W_l z_{l-1}$.
Then we have
\begin{align*}
 \eta :=& \left( W_l z_{l-1} \right)_{k} \\
 =& \sum_{j} (W_l)_{kj} (z_{l-1})_{j} \sim
 \begin{cases}
  {\mathcal{N}}(0, \frac{2}{m} \|z_{l-1}\|_2^2) & l \neq 1 \\
  {\mathcal{N}}(0, \frac{2}{d}) & l = 1
 \end{cases}
\end{align*}

We also define the $k^{\text{th}}$ column vector of $W_{l+1}$:
\[
 W_{l+1}^{(k)} := ((W_{l+1})_{1k}, \ldots ,(W_{l+1})_{mk})^\mathrm{T}
\]

By the assumption, we know that 
\[
 W_{l+1} z_{l} = \eta W_{l+1}^{(k)}.
\]

By the definition of weight initialization of $W$, we have
$\|  W_{l+1}^{(k)} \|_2^2 \sim {\mathcal{N}} (0,2)$.

We define 
$\tilde{z}_{L} := G_L W_{L-1} \cdots G_{l+1} W_{l+1}^{(k)}$.
By the above lemma, we know that with probability at least 
$1- e^{- \Omega(m\varepsilon^2/L)}$, 
\[
 \|\tilde{z}_L \|_2^2 ~\in~ [1-\varepsilon, 1+ \varepsilon].
\]

By definition of $W_{L}$, we have 
$(W_{L})_{ij} \sim {\mathcal{N}} (0, \frac{2}{m})$.

This implies that each element of $W_{L} \tilde{z}_{L}$ is in 
distribution identical to 
${\mathcal{N}} (0, \frac{2}{m} \|\tilde{z}_{L}\|_2^2)$.

Taking into consideration the estimates mentioned above,
we have with probability at least $1- e^{- \Omega(m\varepsilon^2/L)}$:
\[
(M)_{ij} \sim
\begin{cases}
 \mathcal{N}(0,\frac{4}{m^2})\times [1-2\varepsilon,1+3\varepsilon]&l \neq 1 \\
 \mathcal{N}(0, \frac{4}{md}) \times [1-2\varepsilon,1+3\varepsilon]&l = 1
\end{cases}
\]

We complete the proof.

\end{document}